%% file: iclr2027_conference.tex
\documentclass{article} 
\PassOptionsToPackage{table}{xcolor}
\usepackage{iclr2027_conference,times}

\input{math_commands.tex}

\usepackage{hyperref}
\usepackage{url}
\usepackage{booktabs}
\usepackage{multirow}
\usepackage{tabularx}
\usepackage{capt-of}
\usepackage{placeins}
\usepackage{graphicx}
\usepackage{amsmath,amssymb,amsthm}
\usepackage{enumitem}

\graphicspath{{Figures/}}

\title{SocialRL: Refining LLMs' Social Intelligence through Multi-turn Reinforcement Learning and Reward Design}

\iclrfinalcopy

\author{\textbf{Jianing Wang}$^{1}$, \textbf{Xintao Wang}$^{1}$, \textbf{Aili Chen}$^{1}$, \textbf{Jie Shi}$^{1}$, \textbf{Hongcheng Guo}$^{1}$ \\
\textbf{Jun Gao}$^{2}$ \textbf{,} \textbf{Wenxuan Zhao}$^{2}$ \textbf{,} \textbf{Chengkun Lang}$^{2}$ \textbf{,}  \textbf{Yuanli Guo}$^{1}$ \textbf{,} \textbf{Yanghua Xiao}$^{1}$\thanks{Corresponding author.} \\
$^{1}$Fudan University \quad $^{2}$Hello Group \\
\texttt{jnwang26@m.fudan.edu.cn}
}

\begin{document}

\maketitle

\input{Sections/0_abstract}
\input{Sections/1_introduction}
\input{Sections/2_related_work}

\input{Sections/3_method}
\input{Sections/4_experiments}

\input{Sections/5_conclusion}

\bibliography{iclr2027_conference}
\bibliographystyle{iclr2027_conference}

\input{Sections/7_appendix}

\end{document}

%% file: math_commands.tex
\usepackage{amsmath,amsfonts,bm}

\def\eqref#1{equation~\ref{#1}}

\def\1{\bm{1}}

\DeclareMathAlphabet{\mathsfit}{\encodingdefault}{\sfdefault}{m}{sl}
\SetMathAlphabet{\mathsfit}{bold}{\encodingdefault}{\sfdefault}{bx}{n}



%% file: Sections/0_abstract.tex
\begin{abstract}
Social intelligence enables agents to read social context, infer intent, and adapt over sustained dialogue. 
As language models become autonomous collaborators, it is central to building effective and trustworthy human-AI interaction. 
Existing reinforcement learning methods optimize single-turn utterances and sparse outcome rewards, producing short-sighted policies that struggle to manage goal-relationship tensions across multi-turn interactions.
We propose \textbf{SocialRL}, a multi-turn reinforcement learning framework addressing both challenges.
First, we apply multi-turn reinforcement learning using PPO that propagates delayed outcome rewards back to each turn, enabling long-horizon planning.
Second, we design six process reward dimensions capturing the goal-relationship trade-off, including goal advancement, relational attunement, contextual coherence, etc.
A reward model dynamically generates fine-grained scoring criteria for each dimension, while a stage-aware weight schedule prioritizes relationship-building in early turns, goal advancement mid-way, and balanced closure late.
Across multiple social-dialogue benchmarks, SocialRL improves Goal Achievement by an average of $9.2$ percentage points over the corresponding Base models. 
These results demonstrate the effectiveness of SocialRL across synthetic and real social scenes, as well as standard and challenging social scenarios.
The project code is available at \url{https://github.com/wjnwjnwj/SocialRL}.
\end{abstract}

%% file: Sections/1_introduction.tex
\section{Introduction}

\textbf{Social intelligence} is the ability to understand social dynamics, recognize others' intentions, and adjust one's responses throughout extended conversations~\citep{zhou2024sotopia}. 
For humans, social intelligence enables meaningful relationships with friends, family, and colleagues, creating mutual benefit and trust~\citep{stafford1991maintenance}. 
As language models transition from information systems to autonomous agents, social intelligence becomes essential, determining whether AI can collaborate with humans in depth and earn their confidence. 
This skill requires a fundamental tension: pursuing one's goals while preserving relationships with others. 
Social dialogue possesses properties that make it a natural fit for reinforcement learning~\citep{ouyang2022instructgpt}. 
Interactions unfold sequentially, with each turn responding to the present moment while shaping what comes next. 
Meaningful outcomes—goal achievement and relationship quality—emerge only at conversation's end, providing delayed feedback that RL can leverage. 

Existing methods mainly fall into two categories: single-turn reinforcement learning methods and outcome reward training methods. 
Single-turn reinforcement learning methods, such as Sotopia-RL~\citep{yu2025sotopiarl}, optimize individual utterances but neglect cumulative value over multi-turn dialogues. 
Outcome reward training methods, like ArCHer~\citep{zhou2024archer}, use final outcome scores to guide learning, yet provide little intermediate supervision for how each turn should manage social trade-offs. 
However, both remain limited in modeling multi-turn social dynamics: they either favor locally fluent responses without optimizing conversation-level behavior, or rely on sparse outcome feedback without process supervision.
Effective learning requires optimizing multi-turn dialogue trajectories while decomposing social quality into turn-level, multi-dimensional rewards that balance goal progress and relationship health as conversations evolve.

\begin{figure}[t]
\begin{center}
\includegraphics[width=0.98\linewidth]{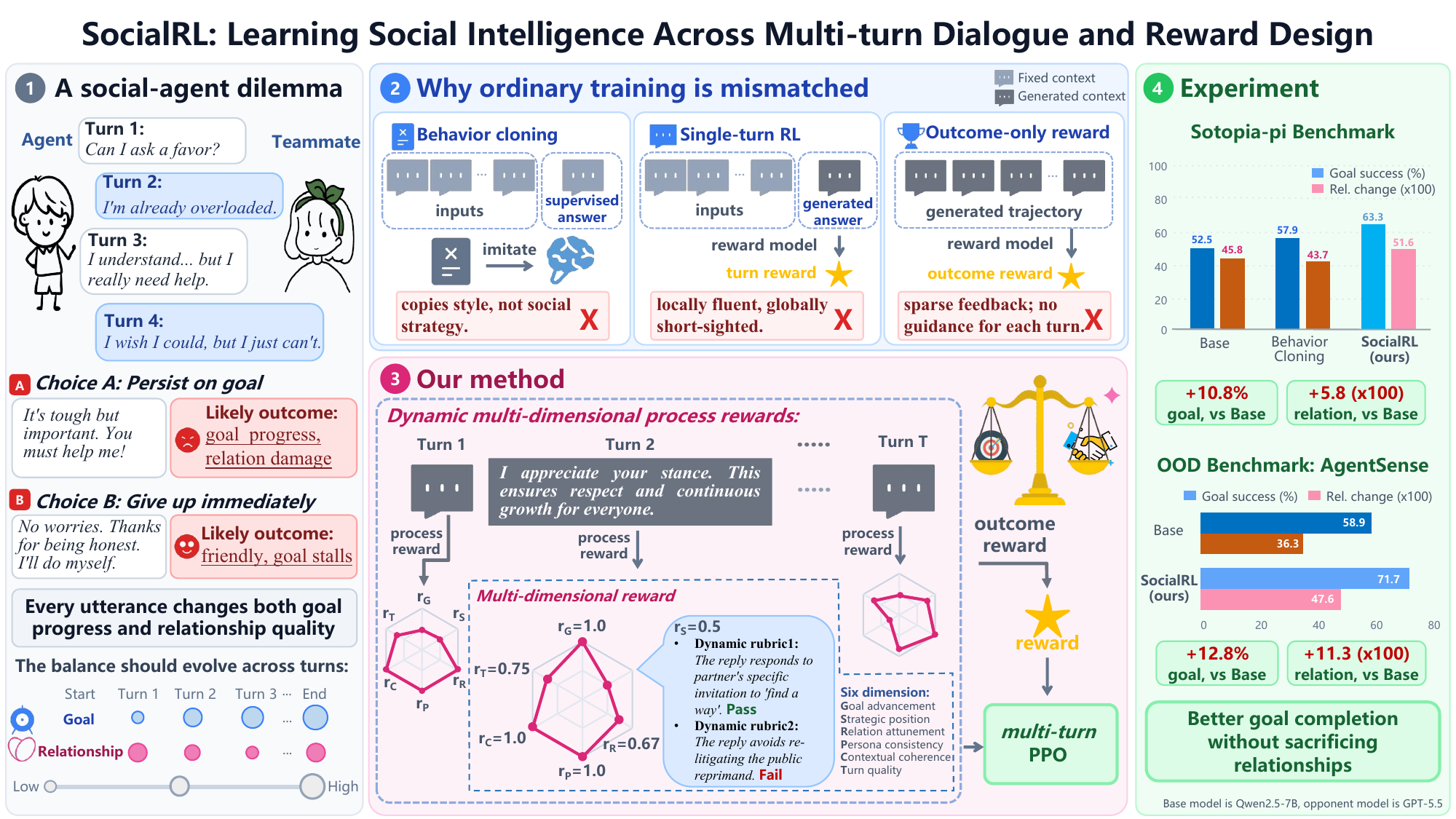}
\end{center}
\caption{Overview of the SocialRL motivation and solution. 
Social dialogue requires agents to balance private goals with relationship management over multi-turn interactions. 
Prior single-turn reinforcement learning or sparse outcome reward methods provide limited long-horizon planning, while SocialRL combines multi-turn reinforcement learning with process rewards.}
\label{fig:problem-overview}
\end{figure}

To address these challenges, we introduce \textbf{SocialRL}, a multi-turn reinforcement learning framework with a multi-dimensional reward system for multi-turn social dialogue, which has two core innovations:

\begin{enumerate}[leftmargin=*]
\item \textbf{Multi-turn trajectory optimization.} 
We treat multi-turn conversation trajectories as the optimization unit and train with PPO online method. 
A value network estimates returns from the turn-level reward sequence, allowing the policy to optimize goal pursuit and relationship maintenance over complete dialogues rather than isolated responses.

\item \textbf{Multi-dimensional dynamic process rewards.} 
We decompose the goal-relationship trade-off into six process reward dimensions: goal advancement and strategic positioning on the goal side~\citep{locke1990goal,kellermann1992communication,berger1997planning}, relational attunement and persona consistency on the relationship side~\citep{stafford1991maintenance,goffman1959presentation}, and contextual coherence and turn quality as enabling conditions for both~\citep{grice1975logic,sacks1974simplest}. 
A reward model generates context-specific scoring criteria for each dimension after every turn, while a stage-aware weight schedule adapts priorities across dialogue phases, prioritizing relationship-building early, goal advancement mid-way, and balanced closure near the end, grounded in classic theories such as Bales's Interaction Process Analysis~\citep{bales1950ipa} and Knapp's relational stage model~\citep{knapp1978relational}.
\end{enumerate}

Large-scale experiments are conducted on SOTOPIA~\citep{zhou2024sotopia}, SOTOPIA-$\pi$~\citep{wang2024sotopiapi}, and AgentSense~\citep{mou2025agentsense}.
After aligning the $[0,10]$ SOTOPIA-All and SOTOPIA-Hard scores to percentage-point units, SocialRL improves Goal Achievement by $9.2$ percentage points on average across the four benchmarks relative to the corresponding Base models.

Our main contributions are as follows. 
(1) We apply multi-turn reinforcement learning to social dialogue, enabling long-horizon planning that balances goal pursuit and relationship management. 
(2) We construct a multi-dimensional process reward system whose six dimensions are grounded in the goal-relationship structure of social dialogue quality, combined with a stage-aware weight schedule that adapts to shifting priorities across dialogue phases. 
(3) Our trained policies outperform the compared non-commercial baselines on SOTOPIA series and AgentSense tasks, while commercial reference models remain an upper bound in several settings.

%% file: Sections/2_related_work.tex
\section{Related Work}

Social intelligence requires language agents to infer intent, track interaction state, and plan long-horizon strategies while balancing private goals with relationship management.

\paragraph{Social Intelligence of LLMs.}
Recent benchmarks instantiate this challenge in interactive social scenarios. 
SOTOPIA~\citep{zhou2024sotopia} and AgentSense~\citep{mou2025agentsense} cover negotiation, empathy, and information verification, while ToMBench~\citep{chen2024tombench} probes theory-of-mind skills; Lifelong-SOTOPIA~\citep{goel2025lifelong} extends evaluation to long-term, multi-scenario relationships. 
These environments provide standardized, human-aligned evaluation, but successful agents must reason over multi-turn trajectories rather than isolated turns.

Beyond benchmark design, a second line of work improves social behavior without online value modeling, via imitation, preference learning, or reward redesign. 
Behavior cloning from the expert trajectories in SOTOPIA-$\pi$~\citep{wang2024sotopiapi} is fluent but limited by data coverage. 
SDPO~\citep{kong2025sdpo} applies segment-level preference optimization, ARIA~\citep{yang2025aria} aggregates rewards in an intention space, and SAVOIR~\citep{feng2026savoir} attributes multi-turn outcomes to utterances via expected utility and Shapley values. 
These give finer supervision, but static demonstrations, preferences, or post-hoc attribution are insufficient for open-ended dialogue whose objectives are multi-dimensional and change in relative importance across stages.

\paragraph{Reinforcement Learning for LLMs.}
Reinforcement learning training~\citep{ouyang2022instructgpt} optimizes interaction outcomes beyond fixed demonstrations. 
Single-turn reinforcement learning methods such as Sotopia-RL~\citep{yu2025sotopiarl}, which applies GRPO~\citep{shao2024grpo}, optimize each utterance independently and thus miss how early actions shape later social outcomes. 
Recent methods extend reinforcement learning or value learning to longer sequences: ArCHer~\citep{zhou2024archer} learns a high-level value function over interaction turns, SVPO~\citep{chen2024svpo} learns step-level preferences and values for mathematical reasoning, REFUEL~\citep{gao2025refuel} regresses relative future returns in multi-turn RLHF, and OMAR~\citep{jiang2026omar} trains conversational agents via multi-agent self-play. 
Yet sparse outcome rewards still give high-variance value estimates, and none explicitly model the stage-dependent goal--relationship trade-off.

In contrast, we jointly optimize full multi-turn trajectories and dense process rewards; as Table~\ref{tab:related-compare} shows, no prior method combines all three.

\input{Tables/tab_related_compare}

%% file: Tables/tab_related_compare.tex
\definecolor{oursRow}{RGB}{231,239,247}
\definecolor{groupRow}{RGB}{244,246,248}
\begin{table*}[t]
\centering
\caption{Comparison of social dialogue optimization methods discussed in Related Work.}
\label{tab:related-compare}
\scriptsize
\setlength{\tabcolsep}{4pt}
\renewcommand{\arraystretch}{0.95}
\begin{tabular}{@{}ll>{\raggedright\arraybackslash}p{4.2cm}>{\raggedright\arraybackslash}p{3.3cm}@{}}
\toprule
Method & Unit & Feedback / reward & Value model \\
\midrule
\rowcolor{groupRow}\multicolumn{4}{@{}l}{\textit{Imitation}} \\
BC & Single-turn & Expert demonstrations & None \\
\midrule
\rowcolor{groupRow}\multicolumn{4}{@{}l}{\textit{Preference / reward design}} \\
SDPO & Segment & Preference pairs & None \\
ARIA & Multi-turn & Intention-space reward aggregation & None \\
SAVOIR & Multi-turn & Expected-utility Shapley attribution & Utterance reward model \\
\midrule
\rowcolor{groupRow}\multicolumn{4}{@{}l}{\textit{Single-turn reinforcement learning}} \\
Sotopia-RL (GRPO) & Single-turn & Reward-model scores & GRPO normalization \\
\midrule
\rowcolor{groupRow}\multicolumn{4}{@{}l}{\textit{Multi-turn reinforcement learning}} \\
ArCHer & Multi-turn & Macro-level returns & Macro-level value function \\
SVPO & Multi-step & Step-level preference pairs & Explicit value model \\
REFUEL & Multi-turn & Relative future-return regression & Relative $Q$ estimation \\
OMAR & Multi-turn & Multi-agent self-play & Value estimates with GAE \\
\midrule
\rowcolor{oursRow}\textbf{SocialRL} & \textbf{Full trajectory} & \textbf{Multi dimension process + outcome rewards} & \textbf{PPO value network} \\
\bottomrule
\end{tabular}
\end{table*}

%% file: Sections/3_method.tex
\section{SocialRL}

\begin{figure}[t]
\begin{center}
\includegraphics[width=0.98\linewidth]{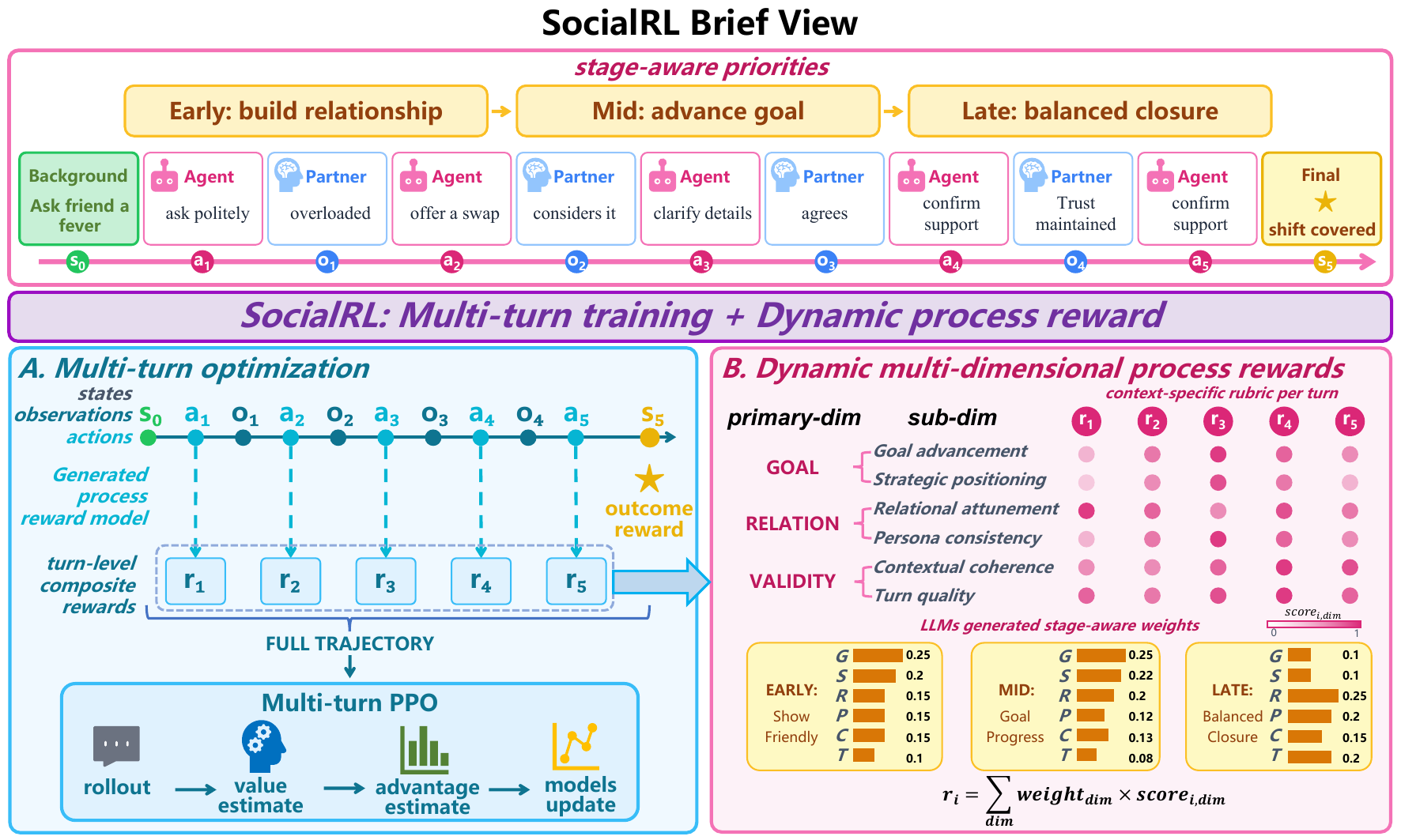}
\end{center}
\caption{SocialRL has two core components. 
Multi-turn trajectory optimization uses a value network, GAE, and PPO to estimate returns from turn-level rewards and optimize complete dialogues. 
Dynamic process reward design scores each policy utterance along six social dimensions, applies context-inferred stage-aware weights, and combines dense process feedback with outcome reward at the final turn.}
\label{fig:method-framework}
\end{figure}

SocialRL trains dialogue policies with multi-turn reinforcement learning. 
We model each interaction as a finite-horizon dialogue MDP, optimize multi-turn trajectories with PPO, and compute dense process rewards from multi-dimensional social feedback with stage-aware weights. 
Figure~\ref{fig:method-framework} summarizes the overall training framework.

\subsection{Task Formulation}
\label{sec:env}

Each task instance is a triple $(S,G,C)$, where $S$ is the scenario, $G$ is the agent's private social goal, and $C$ contains persona information for both parties. 
We model the interaction as a finite-horizon MDP~\citep{sutton2018reinforcement,puterman1994markov} $\mathcal{M} = (\mathcal{S}, \mathcal{A}, P, r, \gamma, \rho_0, M)$.
The state contains $(S,G,C)$ and the dialogue history; actions are tokens grouped into utterances; $P$ appends both the agent utterance and the counterpart response; $\gamma$ is the discount factor, $\rho_0$ is the initial-state distribution, and $M$ is the maximum number of turns. 
Rewards are assigned at the utterance level: tokens within a turn receive zero reward, and the completed utterance receives $r_m \in [-R_{\max},R_{\max}]$.

Thus a trajectory is $\tau = \bigl(s_0,\; \mathbf{a}_0,\; r_1,\; s_1,\; \dots,\; \mathbf{a}_M,\; r_M,\; s_M\bigr)$, with utterance likelihood $\pi_\theta(\mathbf{a}_m \mid s_{m-1}) = \prod_{t=1}^{L_m} \pi_\theta\!\left(a_t^{(m)} \mid s_{m-1}, a_{<t}^{(m)}\right)$, where $L_m$ is the length of utterance $m$, $a_t$ is the $t^{th}$ token of utterance. 
Besides, a trajectory may terminate early after goal completion. 
With the composite reward $r_m=\tilde{R}_m$ defined in Section~\ref{sec:reward}, the policy maximizes $J(\theta) = \mathbb{E}_{\tau \sim \pi_\theta}\bigl[R(\tau)\bigr]$.
The counterpart's private goal is hidden and their responses are stochastic, making conversation-level optimization under evolving dialogue states essential.

\subsection{Training Algorithm}
\label{sec:algo}

SocialRL uses PPO~\citep{schulman2017ppo} over multi-turn dialogue trajectories. 
For each scenario, the current policy interacts with the counterpart for up to $M$ turns; after rollout, each utterance receives the composite reward $\tilde{R}_t$ from Section~\ref{sec:reward}. 
A value network estimates future returns from these turn-level rewards to compute advantages for trajectory optimization.

We compute Generalized Advantage Estimation (GAE)~\citep{schulman2015gae} from turn-level rewards:
\begin{equation}
\delta_t = \tilde{R}_t + \gamma V_\phi(s_{t+1}) - V_\phi(s_t),
\end{equation}
\begin{equation}
\hat{A}_t = \sum_{l=0}^{M-t}(\gamma\lambda)^l\,\delta_{t+l},
\end{equation}
where $V_\phi(s_{t+1})=0$ at the final turn and $\lambda=0.95$. Unlike token-level RLHF, we treat each utterance as one optimization unit: all tokens in the utterance share the same reward and advantage. 
The clipped policy objective is
\begin{equation}
L^{\mathrm{CLIP}}(\theta)=\mathbb{E}_t\Big[\min\big(\rho_t(\theta)\,\hat{A}_t,\;
\mathrm{clip}(\rho_t(\theta),1-\epsilon,1+\epsilon)\,\hat{A}_t\big)\Big],
\end{equation}
where $\rho_t(\theta)=\pi_\theta(a_t\mid s_t)/\pi_{\theta_{\mathrm{old}}}(a_t\mid s_t)$ is the utterance likelihood ratio and $\epsilon=0.2$. 
The value network is trained with
\begin{equation}
\mathcal{L}^V(\phi)=\mathbb{E}_t\big[(V_\phi(s_t)-G_t)^2\big],
\end{equation}
where $G_t=\sum_{l=0}^{M-t}\gamma^l\tilde{R}_{t+l}$ is the discounted return from turn $t$.

We use PPO rather than GRPO~\citep{shao2024grpo} because social dialogue rewards are delayed and strongly state-dependent. 
GRPO normalizes rewards across a group of sampled trajectories and replaces PPO's state baseline with a group-level baseline. 
For turn $m$, PPO uses
\label{sec:ppo-vs-grpo}
\begin{equation}
b_m^{\mathrm{PPO}}=V^\pi(s_{m-1})=\mathbb{E}[G_m\mid s_{m-1}],
\end{equation}
where $G_m=\sum_{k=m}^{M}\gamma^{k-m}r_k$. GRPO instead uses
\begin{equation}
b_m^{\mathrm{GRPO}}=\mu_0D_m, \qquad
D_m=\sum_{k=m}^{M}\gamma^{k-m},
\end{equation}
where $\mu_0$ is the large-group mean of the turn-level rewards. 
GRPO has a finite-group correction that decreases as group size grows, while PPO's state-only baseline leaves the reference gradient target unchanged. 
The key long-horizon difference is variance: under the standard policy-gradient baseline approximation, the unnormalized GRPO estimator adds a mismatch term whenever dialogue states with the same turn index have different values:
\begin{equation}
\widetilde{V}_{\mathrm{GRPO}} \approx V_{\mathrm{PPO}} + \Delta, \qquad
\Delta=\sum_{m=1}^{M}\mathbb{E}\!\left[\|U_m\|^2\bigl(V^\pi(s_{m-1})-\mu_0D_m\bigr)^2\right]\geq 0,
\end{equation}
where $U_m=\nabla_\theta\log\pi_\theta(a_m\mid s_{m-1})$. 
The actual normalized GRPO variance additionally carries the global factor $1/\sigma_0^2$. 
Since the same turn index can correspond to negotiation, repair, compromise, or failure, the baseline-mismatch term becomes more pronounced in longer-horizon dialogues. 
Appendix~\ref{app:theory} provides the derivation.

\subsection{Reward Design}
\label{sec:reward}

Social dialogue requires feedback on both goal progress and relationship management. 
Since outcome-only scores provide no direct supervision for intermediate social behavior over 10--20 turns, SocialRL uses a multi-dimensional process reward and combines it with the outcome reward~\citep{lightman2023verify}. 
At turn $t$, the composite reward is
\begin{equation}
\tilde{R}_t =
\begin{cases}
\alpha \cdot R_{\mathrm{process},t} + \beta \cdot R_{\mathrm{outcome}}, & t = T, \\
\alpha \cdot R_{\mathrm{process},t}, & t < T,
\end{cases}
\end{equation}
where $T$ is the final turn, $R_{\mathrm{process},t}$ scores the current utterance, $R_{\mathrm{outcome}}$ scores final success, and validation selects $\alpha=0.3,\beta=1$. The process reward is
\begin{equation}
R_{\mathrm{process},t}=\sum_{i=1}^{6}w_{t,i}r_{t,i},
\end{equation}
with normalized non-negative weights
\begin{equation}
\sum_{i=1}^{6}w_{t,i}=1, \qquad w_{t,i}\geq 0.
\end{equation}
Intermediate turns therefore still produce useful gradients even when final success is not yet known or the dialogue ultimately fails.

Each utterance is scored along six dimensions: \textit{goal advancement}, \textit{strategic positioning}, \textit{relational attunement}, \textit{persona consistency}, \textit{contextual coherence}, and \textit{turn quality}. 
These dimensions separate goal-side progress, relationship-side maintenance, and general conversational validity, drawing on goal-setting theory~\citep{locke1990goal}, strategic communication~\citep{kellermann1992communication,berger1997planning}, relationship maintenance~\citep{stafford1991maintenance}, self-presentation~\citep{goffman1959presentation}, relevance~\citep{grice1975logic}, and conversation analysis~\citep{sacks1974simplest}.

To make scoring context-specific, we use a rubric mechanism inspired by LLM-based evaluation and criterion-guided feedback~\citep{zheng2023judging,liu2023geval,bai2022constitutional}. 
Given scenario $S$, persona $C$, goal $G$, history $H_t$, and utterance $a_t$, the reward model generates 2--4 binary criteria for each dimension and judges whether each criterion passes. 
The dimension score is the pass rate:
\begin{equation}
r_{t,i}=\frac{1}{M_i}\sum_{j=1}^{M_i}\mathrm{pass}_{i,j},
\end{equation}
where $\mathrm{pass}_{i,j}\in\{0,1\}$. 
Binary criteria are more stable than abstract scalar judgments and make reward feedback inspectable.

Finally, SocialRL changes dimension weights across dialogue stages. 
Following interaction and relational-stage theories~\citep{bales1950ipa,knapp1978relational}, early turns emphasize coherence, persona, and rapport; middle turns emphasize goal advancement and strategy; late turns balance closure with relationship preservation. 
The reward model infers the current stage from context rather than fixed turn-ratio thresholds:
\begin{equation}
\phi_t=f_{\mathrm{stage}}(S,C,G,H_t,a_t), \qquad
\phi_t\in\{\mathrm{Early},\mathrm{Mid},\mathrm{Late}\}.
\end{equation}
Given $\phi_t$, it proposes $\hat{\mathbf{w}}_t$, which is sanitized by whitelist filtering, clipping, caps, fallback priors, overflow redistribution, and normalization before computing $R_{\mathrm{process},t}$.

%% file: Sections/4_experiments.tex
\section{Experiments}

We evaluate SocialRL on four benchmarks with four trained backbones and four opponent models. 
We ask whether it improves social dialogue performance, which design choices account for the gains, and whether it induces long-horizon strategies.

\subsection{Experimental Setup}
\label{sec:setup}

We use two judge-based metrics~\citep{zheng2023judging,liu2023geval}: Goal Achievement and Relationship Change. 
Goal Achievement is reported as a percentage on SOTOPIA-$\pi$ and AgentSense, and as a $[0,10]$ score on SOTOPIA-All and SOTOPIA-Hard; Relationship Change is reported after multiplying the SOTOPIA-$\pi$ and AgentSense values by 100, while the original SOTOPIA benchmarks retain their $[-5,5]$ scale. 
Unless otherwise stated, reported means and standard deviations are computed from five independent repeated experiments of each policy--opponent pair.
Prompts are provided in Appendix~\ref{app:eval-prompts}.

We use three complementary interactive benchmarks. 
SOTOPIA-$\pi$ is constructed from cleaned and filtered synthetic dialogues~\citep{wang2024sotopiapi}; it contains 1{,}773 scenarios across seven social contexts, including 259 held-out test scenarios. 
Its multi-turn interactions evaluate whether an agent can pursue a social goal while maintaining the counterpart's relationship quality. 
SOTOPIA-All and SOTOPIA-Hard are based on the open-ended SOTOPIA environment~\citep{zhou2024sotopia}. 
SOTOPIA-All covers 90 diverse everyday social scenarios, whereas SOTOPIA-Hard is a 14-scenario subset selected for higher conflict, ambiguous intentions, and subtle social norms. 
AgentSense is an independent benchmark built bottom-up from social situations extracted from movie and television scripts~\citep{mou2025agentsense}. 
It provides 1{,}225 scenarios organized into 245 templates with synthetic character instantiations, and evaluates both goal completion and latent social reasoning, such as inferring private information from dialogue.

We train Qwen2.5-7B-Instruct, Qwen3-8B, LLaMA3.1-8B, and Gemma-3-4B policies against Qwen2.5-7B, Qwen3-8B, Qwen3.5-35B, and GPT-5.5 opponents. 
Baselines include base models, Behavior Cloning (BC), Sotopia-RL~\citep{yu2025sotopiarl}, SDPO~\citep{kong2025sdpo}, ArCHer~\citep{zhou2024archer}, and commercial reference models. 
Implementation details are in Appendix~\ref{app:impl}.

\subsection{Main Results}
\label{sec:main}

\paragraph{SocialRL improves goal achievement and relationship maintenance.}
We first evaluate the main two-party setting on SOTOPIA-$\pi$ (Table~\ref{tab:sotopia-pi}). 
The table reports Goal Achievement success rates and Relationship Change across four opponents. 
Across opponents, SocialRL with the Qwen2.5-7B backbone averages $52.3\%$ / $0.358$ in Goal Achievement / Relationship Change, compared with $42.3\%$ / $0.311$ for Base. 
The strongest trained-model average is obtained by the LLaMA3.1-8B policy ($59.6\%$ / $0.341$), while GPT-5.5 reaches $69.0\%$ / $0.388$ as a commercial reference.

\input{Tables/tab_sotopia_pi}

Across opponents, LLaMA3.1-8B has the strongest trained-model average in Goal Achievement ($59.6\%$), while Qwen2.5-7B leads in Relationship Change ($0.358$). 
The smaller Gemma-3-4B backbone also benefits substantially: its average rises from $30.5\% / -0.093$ for Base to $52.3\% / 0.134$ with SocialRL, improving by $21.8$ percentage points in Goal Achievement and $0.227$ in the displayed Relationship Change scale.

As a separate and out of distribution test of multi-party interaction and latent social reasoning, AgentSense results are shown in Table~\ref{tab:agentsense}. 
SocialRL with Qwen3-8B reaches $80.5\% / 0.444$ on average, exceeding the corresponding Base model by $7.2$ percentage points in Goal Achievement and $0.293$ in the displayed Relationship Change scale. 
The improvement is especially clear for Gemma-3-4B, which rises from $46.3\% / 0.065$ to $68.1\% / 0.234$.
These results extend the SOTOPIA-$\pi$ findings to script-derived, multi-party scenarios where success also requires inferring information that is not stated explicitly.
we report SOTOPIA-All and SOTOPIA-Hard in Appendix~\ref{app:additional-results}.

To summarize performance across settings, we use each table's opponent-averaged Goal Achievement and compute the absolute SocialRL--Base difference for the same backbone, then average across the four trained backbones. 
This yields gains of $10.725$ points on SOTOPIA-$\pi$, $9.625$ on AgentSense, $0.9275$ on SOTOPIA-All, and $0.7125$ on SOTOPIA-Hard. 
Because the latter two benchmarks use $[0,10]$ Goal Achievement scores, we multiply their gains by 10 before averaging, which places all four benchmarks on a percentage-point scale. 
The resulting overall improvement is $9.2$ percentage points.

\input{Tables/tab_agentsense}

\subsection{Ablation Experiments}
\label{sec:ablation}

\begin{figure}[t]
\centering
\includegraphics[width=\linewidth]{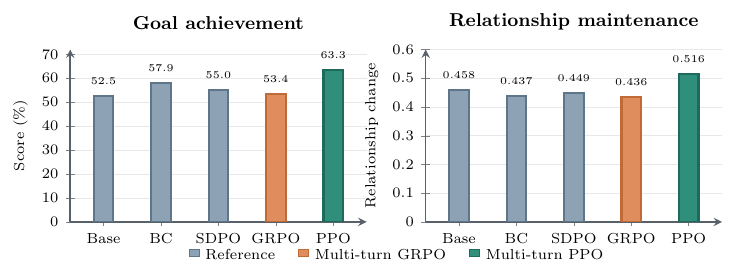}
\caption{Contextual comparison of Qwen2.5-7B policies on the SOTOPIA-$\pi$ benchmark against GPT-5.5. 
Goal values are percentages; Relationship Change is reported on its original $[-1,1]$ scale. 
Base, BC, and SDPO are reference methods; the controlled algorithm comparison uses SocialRL's full reward design with both multi-turn GRPO and multi-turn PPO.}
\label{fig:ablation-algorithm}
\end{figure}

\paragraph{Algorithm choice, process rewards, and dynamic weights matter.}
Figure~\ref{fig:ablation-algorithm} compares Qwen2.5-7B policies against GPT-5.5. 
With the same SocialRL reward design, PPO exceeds multi-turn GRPO by $9.9$ percentage points in Goal Achievement and $0.080$ in Relationship Change. 
The remaining ablations are in Appendix~\ref{app:ablation}: process rewards have the largest effect, goal-oriented dimensions drive task success, relational attunement preserves relationship quality, and dynamic weights outperform fixed schedules.

\subsection{Human Alignment Evaluation}
\label{sec:human-alignment}
To assess whether the automatic reward models agree with human judgments, we conduct an independent human-alignment audit on 100 dialogue trajectories. 
For PRM, the audit covers 497 turn-level process records and yields a correlation of $0.816$ between PRM scores and aggregated human ratings. 
For ORM, the audit covers 100 dialogue-level judgments and yields correlations of $0.930$ for Goal Achievement and $0.946$ for Relationship Change. 
All correlations are computed after the corresponding score transformations and indicate strong agreement between the automatic evaluators and human assessment.

\subsection{In-Depth Case Analysis}
\label{sec:analysis}

\paragraph{SocialRL learns long-horizon social strategy.}
We analyze a representative SOTOPIA-$\pi$ board-game dialogue in which Ethan must speed up Benjamin's play without damaging rapport. 
Representative turns and extended analysis are provided in Appendix~\ref{app:case}.

The Base model is polite but reactive: it repeatedly agrees with Benjamin, who independently proposes the one-minute limit. 
The pacing solution therefore comes from Benjamin rather than Ethan, yielding a goal score of $0.0$.

SocialRL turns the opening into an explicit rule: ``Let's set a quick timer for your next turn,'' and ratifies Benjamin's ``ninety seconds'' proposal with ``Deal. 
Ninety seconds, and no deep dives.'' Benjamin later adopts the rule and requests future enforcement. 
The dialogue reaches a goal score of $1.0$ while preserving Base's relationship delta ($+0.6$).

Thus, SocialRL converts an opening into a face-saving commitment and maintains it across later turns, improving goal success without sacrificing rapport.

%% file: Tables/tab_sotopia_pi.tex
\begin{table}[t]
\caption{%
  Results on the SOTOPIA-$\pi$ benchmark: Goal Achievement (\%) / Relationship Change ($\times100$) for each policy--opponent pair. Values are means across repeated experiments; complete mean $\pm$ SD results are in Appendix~\ref{tab:appendix-sotopia-pi}. Relationship Change values are multiplied by 100.
  Bold denotes the best SocialRL model per metric and opponent column.
  SocialRL has improved performance on each same-backbone network and in the average results among non-commercial methods.
}
\label{tab:sotopia-pi}
\definecolor{baseRow}{RGB}{248,249,250}
\definecolor{baselineRow}{RGB}{242,246,248}
\definecolor{oursRow}{RGB}{231,239,247}
\definecolor{referenceRow}{RGB}{247,241,232}
\begin{center}
\normalsize
\setlength{\tabcolsep}{2pt}
\renewcommand{\arraystretch}{0.92}
\begin{tabular}{lccccc}
\toprule
 & \multicolumn{4}{c}{\textbf{Opponent Model}} & \\
\cmidrule(lr){2-5}
Method & Qwen2.5-7B & Qwen3-8B & Qwen3.5-35B & GPT-5.5 & Avg. \\
\midrule
\rowcolor{baseRow}Base (Qwen2.5-7B)   & 33.5 / 23.5 & 38.4 / 22.8 & 44.6 / 32.1 & 52.5 / 45.8 & 42.3 / 31.1 \\
\rowcolor{baseRow}Base (Qwen3-8B)     & 43.2 / 13.6 & 44.2 / 8.3 & 55.0 / 13.7 & 60.9 / 24.8 & 50.8 / 15.1 \\
\rowcolor{baseRow}Base (LLaMA3.1-8B)  & 44.3 / 21.7 & 47.1 / 19.4 & 55.0 / 17.3 & 63.2 / 37.1 & 52.4 / 23.9 \\
\rowcolor{baseRow}Base (Gemma-3-4B)   & 22.0 / $-$10.6 & 30.3 / $-$6.3 & 32.1 / $-$15.0 & 37.6 / $-$5.4 & 30.5 / $-$9.3 \\
\midrule
\rowcolor{baselineRow}BC (Qwen2.5-7B)     & 38.9 / 27.2 & 43.0 / 23.6 & 49.9 / 22.3 & 57.9 / 43.7 & 47.4 / 29.2 \\
\rowcolor{baselineRow}SDPO (Qwen2.5-7B)   & 33.1 / 24.0 & 42.3 / 23.9 & 45.3 / 31.1 & 55.0 / 44.9 & 43.9 / 31.0 \\
\rowcolor{baselineRow}Sotopia-RL (Qwen2.5-7B) & 41.2 / 26.0 & 43.7 / 20.9 & 49.0 / 17.8 & 60.5 / 39.4 & 48.6 / 26.0 \\
\rowcolor{baselineRow}ArCHer (Qwen2.5-7B) & 31.6 / 22.6 & 39.6 / 22.8 & 46.6 / 31.3 & 48.6 / 41.0 & 41.6 / 29.4 \\
\midrule
\rowcolor{oursRow}SocialRL (Qwen2.5-7B) & 43.3 / 28.2 & 45.8 / 26.9 & 56.7 / \textbf{36.3} & 63.3 / 51.6 & 52.3 / \textbf{35.8} \\
\rowcolor{oursRow}SocialRL (Qwen3-8B) & 47.7 / 18.8 & 48.1 / 20.8 & 61.0 / 18.1 & 62.2 / 24.9 & 54.7 / 20.6 \\
\rowcolor{oursRow}SocialRL (LLaMA3.1-8B) & \textbf{50.2} / \textbf{31.2} & \textbf{54.3} / \textbf{27.6} & \textbf{63.6} / 24.9 & \textbf{70.2} / \textbf{53.0} & \textbf{59.6} / 34.1 \\
\rowcolor{oursRow}SocialRL (Gemma-3-4B) & 42.4 / 10.6 & 48.0 / 6.4 & 58.6 / 13.7 & 60.3 / 22.9 & 52.3 / 13.4 \\
\midrule
\rowcolor{referenceRow}GPT-5.5 (reference) & 57.0 / 31.6 & 64.9 / 32.2 & 75.3 / 40.1 & 78.9 / 51.4 & 69.0 / 38.8 \\
\bottomrule
\end{tabular}
\end{center}
\end{table}

%% file: Tables/tab_agentsense.tex
\definecolor{baseRow}{RGB}{248,249,250}
\definecolor{baselineRow}{RGB}{242,246,248}
\definecolor{oursRow}{RGB}{231,239,247}
\definecolor{referenceRow}{RGB}{247,241,232}
\begin{table}[t]
\caption{%
  AgentSense results: Goal Achievement Success Rate (\%) / Relationship Change ($\times100$) across multi-party scenarios. 
  Bold denotes the best SocialRL model per opponent.
  Values are means across five independent repeats; complete mean $\pm$ SD results are in Appendix~\ref{tab:appendix-agentsense}. Relationship Change values are multiplied by 100. SocialRL consistently outperforms BC, SDPO, and Sotopia-RL on both metrics.
}
\label{tab:agentsense}
\begin{center}
\normalsize
\setlength{\tabcolsep}{2pt}
\renewcommand{\arraystretch}{0.92}
\begin{tabular}{lccccc}
\toprule
 & \multicolumn{4}{c}{\textbf{Opponent Model}} & \\
\cmidrule(lr){2-5}
Method & Qwen2.5-7B & Qwen3-8B & Qwen3.5-35B & GPT-5.5 & Avg. \\
\midrule
\rowcolor{baseRow}Base (Qwen2.5-7B)      & 53.7 / 32.8 & 55.7 / 33.6 & 65.2 / 40.9 & 58.9 / 36.3 & 58.4 / 35.9 \\
\rowcolor{baseRow}Base (Qwen3-8B)        & 67.8 / 33.7 & 72.9 / 38.9 & 69.6 / 35.8 & 82.8 / 48.4 & 73.3 / 39.2 \\
\rowcolor{baseRow}Base (LLaMA3.1-8B)     & 83.2 / 48.1 & 85.2 / 49.2 & 86.2 / 51.9 & 89.4 / 58.1 & 86.0 / 51.8 \\
\rowcolor{baseRow}Base (Gemma-3-4B)      & 44.4 / 6.3 & 45.4 / 6.1 & 48.1 / 9.7 & 47.5 / 3.7 & 46.3 / 6.5 \\
\midrule
\rowcolor{baselineRow}BC (Qwen2.5-7B)        & 60.3 / 36.7 & 56.7 / 32.8 & 63.1 / 37.9 & 67.0 / 46.9 & 61.8 / 38.6 \\
\rowcolor{baselineRow}SDPO (Qwen2.5-7B)      & 57.6 / 35.8 & 58.0 / 36.3 & 63.0 / 37.4 & 66.7 / 48.1 & 61.3 / 39.4 \\
\rowcolor{baselineRow}Sotopia-RL (Qwen2.5-7B) & 63.5 / 37.1 & 59.3 / 33.5 & 63.9 / 35.8 & 69.6 / 49.4 & 64.1 / 38.9 \\
\rowcolor{baselineRow}ArCHer (Qwen2.5-7B)    & 51.1 / 31.9 & 52.6 / 32.0 & 60.9 / 38.9 & 58.9 / 35.7 & 55.9 / 34.6 \\
\midrule
\rowcolor{oursRow}SocialRL (Qwen2.5-7B) & 61.0 / 38.6 & 62.2 / 38.5 & 72.1 / 45.9 & 71.7 / 47.6 & 66.7 / 42.7 \\
\rowcolor{oursRow}SocialRL (Qwen3-8B) & 76.7 / 39.5 & 77.1 / 40.9 & 82.0 / 46.3 & 86.0 / 50.7 & 80.5 / 44.4 \\
\rowcolor{oursRow}SocialRL (LLaMA3.1-8B) & \textbf{84.3} / \textbf{48.3} & \textbf{85.9} / \textbf{50.1} & \textbf{87.6} / \textbf{52.4} & \textbf{90.9} / \textbf{58.9} & \textbf{87.2} / \textbf{52.4} \\
\rowcolor{oursRow}SocialRL (Gemma-3-4B) & 66.7 / 22.5 & 66.4 / 22.3 & 68.8 / 24.7 & 70.4 / 23.9 & 68.1 / 23.4 \\
\midrule
\rowcolor{referenceRow}GPT-5.5 (reference) & 88.1 / 45.6 & 92.2 / 48.9 & 93.8 / 52.6 & 92.5 / 48.9 & 91.7 / 49.0 \\
\bottomrule
\end{tabular}
\end{center}
\end{table}

%% file: Sections/5_conclusion.tex
\section{Conclusion}

We presented SocialRL, a multi-turn reinforcement learning framework for social dialogue that addresses the tension between goal pursuit and relationship management. 
SocialRL optimizes complete dialogue trajectories with PPO and a value network, while its multi-dimensional process reward system provides dense turn-level feedback through rubric-based scoring and stage-aware weights. 
Experiments on the SOTOPIA-$\pi$ benchmark, SOTOPIA-All, SOTOPIA-Hard, and AgentSense show improvements over the compared non-commercial imitation-learning, reinforcement-learning, and preference-learning baselines across multiple opponents and trained backbones; commercial reference models remain stronger in several settings. 
Remaining limitations include over-compromise, rigidity under unexpected opponent moves, and memory decay in very long dialogues, suggesting future work on goal-floor constraints, broader training distributions, opponent modelling, and memory-augmented policies.

%% file: Sections/7_appendix.tex
\newpage
\appendix

\input{Sections/7_appendix/1_theory}
\input{Sections/7_appendix/2_eval_prompts}
\input{Sections/7_appendix/3_impl_details}
\input{Sections/7_appendix/4_extra_benchmarks}

\input{Sections/7_appendix/5_ablation}
\input{Sections/7_appendix/6_extended_case}

%% file: Sections/7_appendix/1_theory.tex
\section{Theoretical Analysis of PPO and GRPO in Multi-Turn Social Dialogue}
\label{app:theory}

We compare PPO with GAE and GRPO for multi-turn social dialogue optimization under the standard policy-gradient baseline approximation.
The argument rests on two properties: gradient bias and gradient variance.

\subsection{Notation}

Table~\ref{tab:notation} summarizes the notation used throughout this appendix and the main text.

\input{Tables/tab_notation}

\subsection{Setup and Notation}

Consider a trajectory with $M$ dialogue turns:
\begin{equation}
   \tau = (s_0, \mathbf{a}_1, r_1, s_1, \dots, \mathbf{a}_M, r_M, s_M).
\end{equation}
Here $s_{m-1}$ is the dialogue state before turn $m$, $\mathbf{a}_m$ is the utterance generated at that turn, and $r_m$ is its reward.
The discounted return from turn $m$ is
\begin{equation}
   G_m = \sum_{k=m}^{M} \gamma^{k-m} r_k,
\end{equation}
where $\gamma\in[0,1]$ controls how strongly future rewards affect turn $m$.
and the score function for the utterance at turn $m$ is
\begin{equation}
   U_m = \nabla_\theta \log \pi_\theta(\mathbf{a}_m \mid s_{m-1}).
\end{equation}
Thus $U_m$ is the gradient direction induced by the sampled utterance.
We use the following trajectory-level Monte Carlo target:
\begin{equation}
   g_{\mathrm{MC}}(\theta)
   =\mathbb{E}\left[\sum_{m=1}^{M} U_m G_m\right].
\end{equation}
$g_{\mathrm{MC}}$ is the expected policy-gradient signal used as the common reference for PPO and GRPO.
All expectations are on-policy.
Assume fixed $M$, sufficient finite moments for the sample-standard-deviation expansion, and $\sigma_0>0$.
Treat the turn index as part of the state and define
\begin{align}
   V^\pi(s)            & = \mathbb{E}[G_m \mid s_{m-1}=s],                          \\
   Q^\pi(s,\mathbf{a}) & = \mathbb{E}[G_m \mid s_{m-1}=s, \mathbf{a}_m=\mathbf{a}], \\
   A^\pi(s,\mathbf{a}) & = Q^\pi(s,\mathbf{a}) - V^\pi(s).
\end{align}
Here $V^\pi$ is the state value, $Q^\pi$ is the state-action value, and $A^\pi$ measures the value of an action relative to the state average.

\textbf{Lemma 1} (Baseline unbiasedness).
For any state-only function
$b(s_{m-1})$,
\begin{equation}
   \hat{g}_b = \sum_{m=1}^M U_m \left(G_m - b(s_{m-1})\right)
\end{equation}
where $b$ is a control-variate baseline used to reduce gradient variance has the same expectation as
\begin{equation}
   \hat{g}_{\mathrm{MC}} = \sum_{m=1}^M U_m G_m.
\end{equation}
\textit{Proof.}
Conditioning on $s_{m-1}$,
\begin{align}
   \mathbb{E}\left[U_m b(s_{m-1})\right]
    & = \mathbb{E}_{s_{m-1}}\left[ b(s_{m-1})\,\mathbb{E}_{\mathbf{a}_m \sim \pi_\theta(\cdot \mid s_{m-1})}\left[\nabla_\theta \log \pi_\theta(\mathbf{a}_m \mid s_{m-1}) \mid s_{m-1}\right] \right] \\
    & = \mathbb{E}_{s_{m-1}}\left[ b(s_{m-1}) \sum_{\mathbf{a}} \pi_\theta(\mathbf{a}\mid s_{m-1}) \nabla_\theta \log \pi_\theta(\mathbf{a}\mid s_{m-1}) \right]                                       \\
    & = \mathbb{E}_{s_{m-1}}\left[ b(s_{m-1}) \sum_{\mathbf{a}} \nabla_\theta \pi_\theta(\mathbf{a}\mid s_{m-1}) \right]                                                                               \\
    & = \mathbb{E}_{s_{m-1}}\left[ b(s_{m-1}) \nabla_\theta \sum_{\mathbf{a}} \pi_\theta(\mathbf{a}\mid s_{m-1}) \right]                                                                               \\
    & = \mathbb{E}_{s_{m-1}}\left[ b(s_{m-1}) \nabla_\theta 1 \right]
   = \mathbf{0}.
\end{align}
Summing over $m$ gives
\begin{equation}
   \mathbb{E}[\hat{g}_b]
   = \mathbb{E}\left[\sum_{m=1}^M U_m G_m\right]
   - \mathbb{E}\left[\sum_{m=1}^M U_m b(s_{m-1})\right]
   = \mathbb{E}[\hat{g}_{\mathrm{MC}}].
\end{equation}
Thus $\mathbb{E}[\hat{g}_b]=g_{\mathrm{MC}}(\theta)$.
The baseline is assumed detached and state-only.
\hfill$\square$

\subsection{PPO with GAE}

PPO uses the GAE estimator~\citep{schulman2015gae} with residual
\begin{equation}
   \delta_m = r_m + \gamma V_\phi(s_m) - V_\phi(s_{m-1}),
\end{equation}
where $V_\phi$ is the learned critic and $\delta_m$ is its one-step TD error.
and advantage
\begin{equation}
   A_m^{\mathrm{GAE}(\lambda)} = \sum_{l=0}^{M-m} (\gamma\lambda)^l \delta_{m+l}.
\end{equation}
The parameter $\lambda$ controls the bias--variance trade-off in multi-step advantage estimation.
For $V_\phi = V^\pi$ and $\lambda=1$,
\begin{align}
   A_m^{\mathrm{GAE}(1)}
    & = \sum_{l=0}^{M-m} \gamma^l \left(r_{m+l} + \gamma V^\pi(s_{m+l}) - V^\pi(s_{m+l-1})\right) \\
    & = \sum_{l=0}^{M-m} \gamma^l r_{m+l} - V^\pi(s_{m-1}) + \gamma^{M-m+1} V^\pi(s_M).
\end{align}
With $V^\pi(s_M)=0$,
\begin{equation}
   A_m^{\mathrm{GAE}(1)} = G_m - V^\pi(s_{m-1}).
\end{equation}
Hence the unclipped on-policy estimator
\begin{equation}
   \hat{g}_{\mathrm{PPO}} = \sum_{m=1}^M U_m\, A_m^{\mathrm{GAE}(\lambda)}
\end{equation}
has expectation $g_{\mathrm{MC}}(\theta)$ for the ideal-value, $\lambda=1$ case.
This does not establish unbiasedness for clipped PPO, an approximate critic, or general $\lambda<1$.
PPO uses
\begin{equation}
   b_m^{\mathrm{PPO}} = V^\pi(s_{m-1}),
\end{equation}
which approximates the score-norm-weighted trace-variance-minimizing baseline~\citep{greensmith2004variance}.

\subsection{GRPO and Its Bias}

GRPO~\citep{shao2024grpo} samples $G$ trajectories from a fixed prompt and normalizes their $M$ turn-level rewards:
\begin{equation}
   \mathcal{R}=\{r_m^{(i)}: i=1,\ldots,G,\; m=1,\ldots,M\}.
\end{equation}
Let $N=GM$ be the number of rewards in the group.
The group mean is
\begin{equation}
   \mu = \frac{1}{N}\sum_{i=1}^{G}\sum_{m=1}^{M} r_m^{(i)}
   = \frac{1}{GM}\sum_{i=1}^{G}\sum_{m=1}^{M} r_m^{(i)},
\end{equation}
and the group standard deviation is
\begin{equation}
   \sigma = \sqrt{\frac{1}{N}\sum_{i=1}^{G}\sum_{m=1}^{M}\left(r_m^{(i)}-\mu\right)^2}.
\end{equation}
Thus $\mu$ and $\sigma$ are the empirical mean and standard deviation shared by all trajectories and turns in the group.
Standardize each reward by these statistics:
\begin{equation}
   \tilde{r}_m^{(i)} = \frac{r_m^{(i)}-\mu}{\sigma}.
\end{equation}
$\tilde r_m^{(i)}$ is the centered and normalized reward used by GRPO.

The multi-turn GRPO advantage is the discounted future sum:
\begin{equation}
   A_m^{\mathrm{GRPO},(i)}
   = \tilde{r}_m^{(i)} + \gamma \tilde{r}_{m+1}^{(i)} + \gamma^2 \tilde{r}_{m+2}^{(i)} + \cdots + \gamma^{M-m}\tilde{r}_M^{(i)}.
\end{equation}
\begin{equation}
   A_m^{\mathrm{GRPO},(i)}
   = \sum_{k=m}^{M}\gamma^{k-m}\tilde{r}_k^{(i)}.
\end{equation}
Substitution gives
\begin{align}
   A_m^{\mathrm{GRPO},(i)}
    & = \sum_{k=m}^{M}\gamma^{k-m}\left(\frac{r_k^{(i)}-\mu}{\sigma}\right)                               \\
    & = \frac{1}{\sigma}\sum_{k=m}^{M}\gamma^{k-m}\left(r_k^{(i)}-\mu\right)                              \\
    & = \frac{1}{\sigma}\left(\sum_{k=m}^{M}\gamma^{k-m}r_k^{(i)} - \mu\sum_{k=m}^{M}\gamma^{k-m}\right).
\end{align}
With
\begin{equation}
   G_m^{(i)} = \sum_{k=m}^{M}\gamma^{k-m}r_k^{(i)}
\end{equation}
and
\begin{equation}
   D_m = \sum_{k=m}^{M}\gamma^{k-m},
\end{equation}
$D_m$ is the total discount mass remaining after turn $m$.
the advantage is
\begin{equation}
   A_m^{\mathrm{GRPO},(i)} = \frac{1}{\sigma}\left(G_m^{(i)} - \mu D_m\right),
\end{equation}
where $D_m$ is the remaining discount mass.
In the large-group limit,
\begin{equation}
   \mu_0=\frac{1}{M}\sum_{k=1}^{M}\mathbb{E}[r_k],
   \qquad
   \sigma_0^2=\frac{1}{M}\sum_{k=1}^{M}\mathbb{E}\!\left[(r_k-\mu_0)^2\right].
\end{equation}
The constants $\mu_0$ and $\sigma_0$ are the population limits of the group mean and standard deviation as $G\to\infty$.

\textbf{Theorem 1} (Finite-group correction for GRPO).
Assume fixed $M$, $\sigma_0>0$, independent trajectory-level samples, and sufficient moments for a delta-method expansion of the sample standard deviation.
Define
\begin{equation}
   \hat{g}_{\mathrm{GRPO}}^{\mathrm{adj}}
   =\sigma_0\hat{g}_{\mathrm{GRPO}}.
\end{equation}
$\hat g_{\mathrm{GRPO}}^{\mathrm{adj}}$ removes GRPO's global normalization
scale so its direction and variance can be compared with the MC target.
Then
\begin{equation}
   \mathbb{E}[\hat{g}_{\mathrm{GRPO}}^{\mathrm{adj}}]
   =g_{\mathrm{MC}}(\theta)+\mathcal{O}(1/G).
\end{equation}
The unadjusted estimator satisfies
\begin{equation}
   \mathbb{E}[\hat{g}_{\mathrm{GRPO}}]
   =\frac{1}{\sigma_0}g_{\mathrm{MC}}(\theta)+\mathcal{O}(1/G).
\end{equation}
The first term is a global positive rescaling induced by reward
standardization, not a change in gradient direction.
It can be absorbed into the learning rate by setting $\eta_{\mathrm{GRPO}}=\sigma_0\eta$.
After this adjustment, the remaining finite-group discrepancy is $\mathcal{O}(1/G)$ and vanishes as $G$ grows.
Here $\sigma_0 = \lim_{G\to\infty}\sigma$.

\textit{Proof.} Since $\sigma$ is estimated from the group, let
\begin{equation}
   \sigma_0 = \lim_{G\to\infty} \sigma
\end{equation}
Keeping $1/\sigma$ explicit,
\begin{equation}
   \hat{g}_{\mathrm{GRPO}}
   = \frac{1}{G}\sum_{i=1}^{G}\sum_{m=1}^{M} U_m^{(i)} A_m^{\mathrm{GRPO},(i)}.
\end{equation}
Substituting the GRPO advantage gives
\begin{align}
   \hat{g}_{\mathrm{GRPO}}
    & = \frac{1}{G\sigma}\sum_{i,m} U_m^{(i)}\left(G_m^{(i)} - \mu D_m\right)   \\
    & = \frac{1}{\sigma}\left(T_1 - T_2\right), \label{eq:grpo-t-decomposition}
\end{align}
where
\begin{equation}
   T_1 = \frac{1}{G}\sum_{i,m} U_m^{(i)} G_m^{(i)}, \quad T_2 = \frac{1}{G^2 M}\sum_{i,j,m,l} D_m U_m^{(i)} r_l^{(j)}.
\end{equation}
$T_1$ is the group-averaged MC gradient term; $T_2$ is the correction induced by reusing the sampled group mean as a baseline.
For the first term,
\begin{equation}
   \mathbb{E}[T_1]
   = \mathbb{E}\left[\sum_{m=1}^{M} U_m G_m\right]
   = g_{\mathrm{MC}}(\theta).
   \label{eq:grpo-t1-expectation}
\end{equation}
For $T_2$, separate cross- and same-trajectory terms:
\begin{equation}
   \mathbb{E}[T_2]
   = \frac{1}{G^2M}\sum_{i\neq j}\sum_{m,l}D_m\mathbb{E}[U_m^{(i)}r_l^{(j)}]
   + \frac{1}{G^2M}\sum_{i=j}\sum_{m,l}D_m\mathbb{E}[U_m^{(i)}r_l^{(i)}].
\end{equation}
For $i\neq j$,
\begin{equation}
   \mathbb{E}[U_m^{(i)}r_l^{(j)}]
   = \mathbb{E}[U_m^{(i)}]\mathbb{E}[r_l^{(j)}].
\end{equation}
Since
\begin{equation}
   \mathbb{E}[U_m^{(i)}]=\mathbf{0},
\end{equation}
all cross-trajectory terms vanish, leaving
\begin{equation}
   \mathbb{E}[T_2]
   = \frac{1}{G^2M}\cdot G\sum_{m=1}^{M}\sum_{l=1}^{M}D_m\mathbb{E}[U_m r_l]
   = \frac{1}{GM}\sum_{m=1}^{M}\sum_{l=1}^{M}D_m\mathbb{E}[U_m r_l].
   \label{eq:grpo-t2-expectation}
\end{equation}
Expand $1/\sigma$ around $\sigma_0$:
\begin{align}
   \frac{1}{\sigma}
    & = \frac{1}{\sigma_0} \cdot \frac{1}{1 + \frac{\sigma - \sigma_0}{\sigma_0}}                                                    \\
    & = \frac{1}{\sigma_0} \cdot \left(1 - \frac{\sigma - \sigma_0}{\sigma_0} + \mathcal{O}\left((\sigma - \sigma_0)^2\right)\right) \\
    & = \frac{1}{\sigma_0} - \frac{\sigma-
      \sigma_0}{\sigma_0^2} + \mathcal{O}\left((\sigma-
   \sigma_0)^2\right). \label{eq:inverse-sigma-expansion}
\end{align}
Substituting Eqs.~\eqref{eq:grpo-t1-expectation}, \eqref{eq:grpo-t2-expectation}, and \eqref{eq:inverse-sigma-expansion} into Eq.~\eqref{eq:grpo-t-decomposition} yields the following asymptotic form.
\begin{align}
   \mathbb{E}[\hat{g}_{\mathrm{GRPO}}]
    & = \frac{1}{\sigma}\left(\mathbb{E}[T_1]-\mathbb{E}[T_2]\right) \\
    & = \frac{1}{\sigma_0}g_{\mathrm{MC}}(\theta)
   - \frac{1}{GM\sigma_0}\sum_{m=1}^{M}\sum_{l=1}^{M}
   D_m\mathbb{E}[U_m r_l] + \mathcal{O}(1/G)                         \\
    & = \frac{1}{\sigma_0}g_{\mathrm{MC}}(\theta)+\mathcal{O}(1/G).
\end{align}
The remainder includes the same-trajectory correction and the fluctuation of
$\sigma$.
Multiplying by $\sigma_0$ gives the adjusted result. \hfill$\square$

\textbf{Corollary.} As $G\to\infty$, GRPO recovers the policy-gradient direction up to the positive scale $1/\sigma_0$; the adjusted estimator has finite-group correction $\mathcal{O}(1/G)$.
Hence finite-group bias diminishes with group size.
The long-horizon difference is governed primarily by variance.

\subsection{Variance Comparison}

For a fair comparison, use PPO and the scale-adjusted large-group GRPO
estimator:
\begin{align}
   g_{\mathrm{PPO}}
    & =\sum_{m=1}^{M}U_m\left(G_m-V^\pi(s_{m-1})\right), \\
   g_{\mathrm{GRPO}}^{\mathrm{adj}}
    & =\sum_{m=1}^{M}U_m\left(G_m-\mu_0D_m\right).
\end{align}
Both have expectation $g_{\mathrm{MC}}(\theta)$. Define
\begin{align}
   \Sigma_{\mathrm{PPO}} & =\mathrm{Cov}(g_{\mathrm{PPO}}),                     &
   V_{\mathrm{PPO}}      & =\mathrm{Tr}(\Sigma_{\mathrm{PPO}}),                   \\
   \Sigma_{\mathrm{GRPO}}^{\mathrm{adj}}
                         & =\mathrm{Cov}(g_{\mathrm{GRPO}}^{\mathrm{adj}}),     &
   V_{\mathrm{GRPO}}^{\mathrm{adj}}
                         & =\mathrm{Tr}(\Sigma_{\mathrm{GRPO}}^{\mathrm{adj}}).
\end{align}
Here $\Sigma$ denotes the gradient covariance matrix, while $V$ is its trace,
i.e., the total gradient-noise power across parameter dimensions.
Define the gradient signal-to-noise ratios
\begin{equation}
   B_{\mathrm{PPO}}
   =\frac{\|g_{\mathrm{MC}}(\theta)\|^2}{V_{\mathrm{PPO}}},
   \qquad
   B_{\mathrm{GRPO}}
   =\frac{\|g_{\mathrm{MC}}(\theta)\|^2}
   {V_{\mathrm{GRPO}}^{\mathrm{adj}}}.
   \label{eq:gradient-snr}
\end{equation}
$B$ measures gradient signal power relative to total gradient noise; larger
values indicate a more reliable update direction.
The scale adjustment is necessary because unadjusted GRPO has signal
$g_{\mathrm{MC}}/\sigma_0$. It does not change the signal-to-noise ratio,
because both signal power and variance scale by $1/\sigma_0^2$.

We use two approximations for an explicit decomposition. For any state-only
baseline $b_m$, assume
\begin{align}
   \mathrm{Cov}\!\left(U_m(G_m-b_m),U_n(G_n-b_n)\right)               & \approx0,
                                                                      &           & m\neq n, \tag{A} \\
   \mathbb{E}\!\left[\|U_m\|^2(G_m-V^\pi(s_{m-1}))\mid s_{m-1}\right] & \approx0.
   \tag{B}
\end{align}
These assumptions are used only for the variance decomposition.

\paragraph{PPO Variance.}
For the turn-level estimator
\begin{equation}
   g_m(b)=U_m(G_m-b_m),
\end{equation}
$g_m(b)$ is the gradient contribution from turn $m$ under baseline $b_m$.
the trace-variance-minimizing scalar baseline given $s_{m-1}=s$ is
\begin{equation}
   b_m^*(s)=\frac{\mathbb{E}\left[\|U_m\|^2G_m\mid s_{m-1}=s\right]}{\mathbb{E}\left[\|U_m\|^2\mid s_{m-1}=s\right]}.
\end{equation}
$b_m^*$ is the scalar state-only baseline minimizing the trace variance of
$g_m(b)$.
Under conditional score-norm independence, it reduces to
\begin{equation}
   b_m^*(s)\approx \mathbb{E}[G_m\mid s_{m-1}=s]=V^\pi(s).
\end{equation}
PPO uses $b_m^{\mathrm{PPO}}=V^\pi(s_{m-1})$. Define
\begin{equation}
   q_m^{\mathrm{PPO}}
   =\mathrm{Tr}\!\left(\mathrm{Cov}\!\left(
   U_m(G_m-V^\pi(s_{m-1}))\right)\right).
\end{equation}
$q_m^{\mathrm{PPO}}$ is PPO's trace-variance contribution at turn $m$.
Under Assumption A, its total variance is
\begin{equation}
   V_{\mathrm{PPO}}\approx\sum_{m=1}^{M}q_m^{\mathrm{PPO}}.
\end{equation}

\paragraph{GRPO Variance.}
For large $G$, GRPO uses
\begin{equation}
   b_m^{\mathrm{GRPO}}=\mu_0D_m.
\end{equation}
Its mismatch from the state value is
\begin{equation}
   \delta V_m=V^\pi(s_{m-1})-\mu_0D_m.
\end{equation}
$\delta V_m$ measures how far GRPO's state-independent group baseline is from
the state-dependent value baseline.
Using
\begin{equation}
   G_m-\mu_0D_m
   =\left(G_m-V^\pi(s_{m-1})\right)+\delta V_m,
\end{equation}
the conditional cross term vanishes under Assumption B. Since
$\mathbb{E}[U_m\delta V_m]=0$, PPO and unnormalized GRPO have the same
per-turn mean. Hence
\begin{equation}
   q_m^{\mathrm{GRPO}}
   \approx q_m^{\mathrm{PPO}}
   +\mathbb{E}\!\left[\|U_m\|^2\delta V_m^2\right].
\end{equation}
$q_m^{\mathrm{GRPO}}$ is the corresponding unnormalized GRPO variance
contribution.
Under Assumption A,
\begin{equation}
   \widetilde{V}_{\mathrm{GRPO}}
   \approx\sum_{m=1}^{M}q_m^{\mathrm{GRPO}}.
\end{equation}
$\widetilde V_{\mathrm{GRPO}}$ denotes GRPO variance before division by the
global normalization factor $\sigma_0^2$.
The normalized estimator satisfies
\begin{equation}
   V_{\mathrm{GRPO}}\approx
   \frac{\widetilde{V}_{\mathrm{GRPO}}}{\sigma_0^2}.
\end{equation}

\paragraph{Variance Comparison.}
Define the accumulated baseline mismatch
\begin{equation}
   \Delta=\sum_{m=1}^{M}
   \mathbb{E}\!\left[\|U_m\|^2\delta V_m^2\right]\geq0.
\end{equation}
$\Delta$ is the total additional variance caused by GRPO's baseline mismatch.
\textbf{Theorem 2} (Baseline-induced variance gap). Under the stated
approximation,
\begin{equation}
   \widetilde{V}_{\mathrm{GRPO}} \approx V_{\mathrm{PPO}} + \Delta,
   \qquad \Delta \geq 0.
\end{equation}
\textit{Proof.} Summing the per-turn relation for $q_m^{\mathrm{GRPO}}$ under
Assumption~A gives
\begin{align}
   \widetilde{V}_{\mathrm{GRPO}}
    & \approx\sum_{m=1}^{M}q_m^{\mathrm{GRPO}}                   \\
    & \approx\sum_{m=1}^{M}q_m^{\mathrm{PPO}}
   +\sum_{m=1}^{M}\mathbb{E}\!\left[\|U_m\|^2\delta V_m^2\right] \\
    & \approx V_{\mathrm{PPO}}+\Delta.
\end{align}
Since every term in $\Delta$ is non-negative, $\Delta\geq0$, with equality iff
$\delta V_m=0$ almost surely wherever $\|U_m\|>0$. \hfill$\square$

For normalized GRPO,
\begin{equation}
   V_{\mathrm{GRPO}} \approx \frac{1}{\sigma_0^2}
   \left(V_{\mathrm{PPO}}+\Delta\right).
\end{equation}
Thus equality holds iff $V^\pi(s_{m-1})=\mu_0D_m$ almost surely on states with
nonzero score norm, for every $m$.

Substituting Theorem~2 into Eq.~\eqref{eq:gradient-snr} gives
\begin{equation}
   \frac{B_{\mathrm{GRPO}}}{B_{\mathrm{PPO}}}
   \approx \frac{V_{\mathrm{PPO}}}
   {V_{\mathrm{PPO}}+\Delta}
   =\frac{1}{1+\Delta/V_{\mathrm{PPO}}}\leq1.
   \label{eq:snr-ratio}
\end{equation}
The inequality is strict exactly when $\Delta>0$. Thus GRPO has lower
gradient signal-to-noise ratio whenever its group-level baseline fails to
match the state value.

\subsection{Effect of Dialogue Horizon on Variance}

We now analyze how the horizon $M$ affects gradient variance.
Since $1/\sigma_0$ is a global scale that can be absorbed into the learning rate, we compare PPO with the scale-adjusted GRPO estimator.
Define the per-turn terms
\begin{align}
   q_m^{(M)}
    & =\mathrm{Tr}\!\left(\mathrm{Cov}\!\left(
   U_m(G_m-V^\pi(s_{m-1}))\right)\right),                \\
   d_m^{(M)}
    & = \mathbb{E}\!\left[\|U_m\|^2 \delta V_m^2\right].
\end{align}
The superscript $(M)$ emphasizes that both the return distribution and the baseline mismatch depend on the dialogue horizon.
Here $q_m^{(M)}$ is the intrinsic PPO variance at turn $m$, and $d_m^{(M)}$ is GRPO's additional baseline-mismatch contribution.
Then
\begin{align}
   V_{\mathrm{PPO},M}
    & \approx \sum_{m=1}^{M}q_m^{(M)},                       \\
   V_{\mathrm{GRPO},M}^{\mathrm{adj}}
    & \approx \sum_{m=1}^{M}\left(q_m^{(M)}+d_m^{(M)}\right)
   =V_{\mathrm{PPO},M}+\Delta_M,
\end{align}
where
\begin{equation}
   \Delta_M=\sum_{m=1}^{M}d_m^{(M)}\geq0.
\end{equation}
$\Delta_M$ is the accumulated GRPO variance gap for an $M$-turn trajectory.
Equivalently, with
\begin{equation}
   \bar q_M=\frac{1}{M}\sum_{m=1}^{M}q_m^{(M)},
   \qquad
   \bar d_M=\frac{1}{M}\sum_{m=1}^{M}d_m^{(M)},
\end{equation}
$\bar q_M$ and $\bar d_M$ are the average intrinsic variance and average
baseline mismatch per turn, respectively. Therefore,
\begin{equation}
   V_{\mathrm{PPO},M}\approx M\bar q_M,
   \qquad
   V_{\mathrm{GRPO},M}^{\mathrm{adj}}
   \approx M(\bar q_M+\bar d_M).
\end{equation}
Thus, if $\bar q_M$ remains bounded away from zero, PPO variance grows with
$M$. If $\bar d_M$ also remains bounded away from zero, GRPO accumulates an
additional variance gap $\Delta_M=\Omega(M)$. This condition is natural in
social dialogue: states at the same turn index may represent acceptance,
rejection, negotiation, repair, or failure. PPO conditions on these states,
whereas $\mu_0D_m$ depends only on the global reward mean and remaining
discount mass. Writing the limiting reward standard deviation at horizon $M$
as $\sigma_{0,M}$, the normalized GRPO variance is
\begin{equation}
   V_{\mathrm{GRPO},M}
   \approx \frac{M(\bar q_M+\bar d_M)}{\sigma_{0,M}^2}.
\end{equation}
$\sigma_{0,M}$ is the population reward standard deviation for trajectories
with horizon $M$.
If $\sigma_{0,M}$ is uniformly bounded above and away from zero, normalization
does not change the horizon order.
The relative gap is
\begin{equation}
   \frac{V_{\mathrm{GRPO},M}^{\mathrm{adj}}}
   {V_{\mathrm{PPO},M}}
   \approx 1+\frac{\bar d_M}{\bar q_M},
   \qquad
   \frac{B_{\mathrm{GRPO},M}}{B_{\mathrm{PPO},M}}
   \approx\frac{1}{1+\bar d_M/\bar q_M}.
\end{equation}
Therefore, $M$ increases both absolute variances under the conditions above,
but the relative gap need not be monotone: it grows, remains constant, or
shrinks according to $\bar d_M/\bar q_M$.

\subsection{Imperfect Value Network}

For an imperfect value network, write
\begin{equation}
   \hat{V}(s) = V^\pi(s) + \epsilon(s),
\end{equation}
$\epsilon(s)$ is the critic's state-dependent value-estimation error. Its mean
squared error is
\begin{equation}
   L = \mathbb{E}[\epsilon(s)^2].
\end{equation}
$L$ averages critic error over the state distribution.
The additional variance over the ideal PPO baseline is
\begin{equation}
   \Delta_\epsilon = \sum_{m=1}^{M}\mathbb{E}\!\left[\|U_m\|^2\epsilon(s_{m-1})^2\right].
\end{equation}
$\Delta_\epsilon$ is the additional PPO variance induced by critic error.
For the detached Monte Carlo baseline, Lemma~1 still applies. Approximate
GAE with $\lambda<1$ may introduce additional bias, which is not analyzed.
The scale-adjusted comparison favors GRPO only if
\begin{equation}
   \Delta_\epsilon > \Delta,
\end{equation}
For normalized GRPO, the condition is
\begin{equation}
   \frac{V_{\mathrm{PPO}}+\Delta}{\sigma_0^2}
   < V_{\mathrm{PPO}}+\Delta_\epsilon.
\end{equation}
Social dialogue produces sharp value changes across negotiation state, repair,
commitment history, and failure, making $\Delta$ potentially large. The
practical comparison also depends on critic error and $\sigma_0$.

\subsection{Summary}

Table~\ref{tab:ppo-grpo-theory} summarizes the comparison.

\input{Tables/tab_ppo_grpo_theory}

GRPO's finite-group correction decreases as $\mathcal{O}(1/G)$. The main
long-horizon difference is variance: PPO accumulates intrinsic per-turn
variance, whereas GRPO additionally accumulates the state-value mismatch
$\Delta_M$.

%% file: Tables/tab_notation.tex
\begin{table*}[h]
\centering
\caption{Notation used in the theoretical analysis and the main text.}
\label{tab:notation}
\small
\begin{tabular}{ll}
\toprule
Symbol & Description \\
\midrule
$\tau$ & Multi-turn dialogue trajectory $\tau=(s_0,\mathbf{a}_1,r_1,\dots,\mathbf{a}_M,r_M,s_M)$ \\
$M$ & Number of dialogue turns in a trajectory \\
$s_{m-1}$ & Dialogue state before turn $m$ \\
$\mathbf{a}_m$ & Utterance (action) generated at turn $m$ \\
$r_m$ & Reward received at turn $m$ \\
$\gamma$ & Discount factor, $\gamma\in[0,1]$ \\
$G_m$ & Discounted return from turn $m$, $G_m=\sum_{k=m}^{M}\gamma^{k-m}r_k$ \\
$\pi_\theta$ & Policy (LLM) with parameters $\theta$ \\
$U_m$ & Score function $\nabla_\theta\log\pi_\theta(\mathbf{a}_m\mid s_{m-1})$ \\
$g_{\mathrm{MC}}(\theta)$ & Monte Carlo policy-gradient target $\mathbb{E}[\sum_{m=1}^{M}U_mG_m]$ \\
$V^\pi(s)$ & State-value function $\mathbb{E}[G_m\mid s_{m-1}=s]$ \\
$Q^\pi(s,\mathbf{a})$ & State-action value function $\mathbb{E}[G_m\mid s_{m-1}=s,\mathbf{a}_m=\mathbf{a}]$ \\
$A^\pi(s,\mathbf{a})$ & Advantage function $Q^\pi(s,\mathbf{a})-V^\pi(s)$ \\
$V_\phi$ & Learned critic (value network) \\
$\delta_m$ & One-step TD error $r_m+\gamma V_\phi(s_m)-V_\phi(s_{m-1})$ \\
$A_m^{\mathrm{GAE}(\lambda)}$ & GAE advantage at turn $m$ \\
$\lambda$ & GAE bias--variance trade-off parameter, $\lambda\in[0,1]$ \\
$b_m$ & State-only control-variate baseline for variance reduction \\
$G$ & Number of sampled trajectories per prompt (group size) \\
$N$ & Number of rewards in a group, $N=GM$ \\
$\mu$ & Empirical group reward mean \\
$\sigma$ & Empirical group reward standard deviation \\
$\tilde{r}_m^{(i)}$ & Standardized reward $(r_m^{(i)}-\mu)/\sigma$ \\
$D_m$ & Remaining discount mass after turn $m$, $D_m=\sum_{k=m}^{M}\gamma^{k-m}$ \\
$\mu_0,\ \sigma_0$ & Population limits of $\mu$ and $\sigma$ as $G\to\infty$ \\
$\Sigma$ & Gradient covariance matrix \\
$V$ & Trace of $\Sigma$ (total gradient-noise power) \\
$B$ & Gradient signal-to-noise ratio \\
$q_m$ & Per-turn trace-variance contribution of the gradient estimator \\
$\delta V_m$ & GRPO baseline mismatch $V^\pi(s_{m-1})-\mu_0D_m$ \\
$\Delta$ & Accumulated variance gap $\sum_{m=1}^{M}\mathbb{E}\left[\lVert U_m\rVert^2\delta V_m^2\right]$ \\
$\epsilon(s)$ & Critic value-estimation error, $\hat{V}(s)=V^\pi(s)+\epsilon(s)$ \\
$L$ & Mean squared critic error $\mathbb{E}[\epsilon(s)^2]$ \\
$\Delta_\epsilon$ & Additional PPO variance induced by critic error \\
$\eta$ & Learning rate \\
\bottomrule
\end{tabular}
\end{table*}

%% file: Tables/tab_ppo_grpo_theory.tex
\begin{center}
\small
\setlength{\tabcolsep}{4pt}
\renewcommand{\arraystretch}{0.95}
\captionof{table}{PPO vs. GRPO for multi-turn social dialogue.}
\label{tab:ppo-grpo-theory}
\begin{tabular}{@{}lcc@{}}
\toprule
Property & PPO (GAE) & GRPO \\
\midrule
Gradient bias & Zero for ideal unclipped PPO & Scale $1/\sigma_0$ $+\mathcal{O}(1/G)$ \\
Gradient variance & $V_{\mathrm{PPO}}$ & $\left(V_{\mathrm{PPO}}+\Delta\right)/\sigma_0^2$ \\
Gradient SNR (scale-adjusted) & $B_{\mathrm{PPO}}$ & $B_{\mathrm{PPO}}/(1+\Delta/V_{\mathrm{PPO}})$ \\
Baseline type & State-dependent $V^\pi(s)$ & Global mean scaled by $D_m$ \\
Return baseline & Per-state value estimate & Group-normalized reward mean \\
Requires value network & Yes & No \\
\bottomrule
\end{tabular}%
\end{center}

%% file: Sections/7_appendix/2_eval_prompts.tex
\section{Prompt Templates}
\label{app:eval-prompts}

\begin{figure}[ht]
\centering
\includegraphics[width=0.96\linewidth]{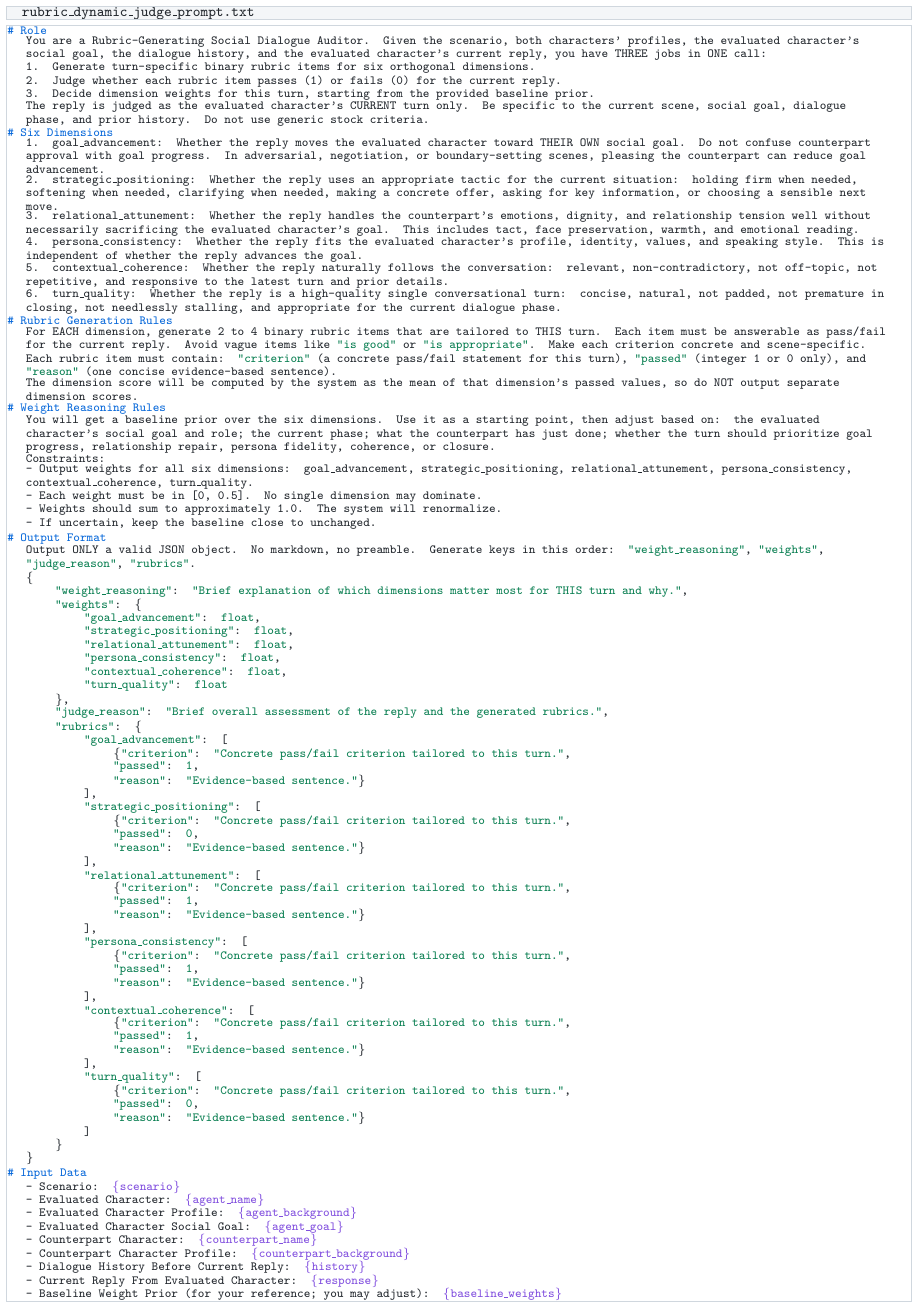}
\caption{Process reward model prompt for turn-level rubric generation, pass/fail
judgment, and context-dependent reward weighting.}
\label{fig:prm-prompt}
\end{figure}

The process reward prompt (Figure~\ref{fig:prm-prompt}) generates turn-specific rubric items for six reward dimensions, judges each item as pass or fail, and assigns context-dependent dimension weights from a baseline prior.

\begin{figure}[ht]
\centering
\includegraphics[width=0.96\linewidth]{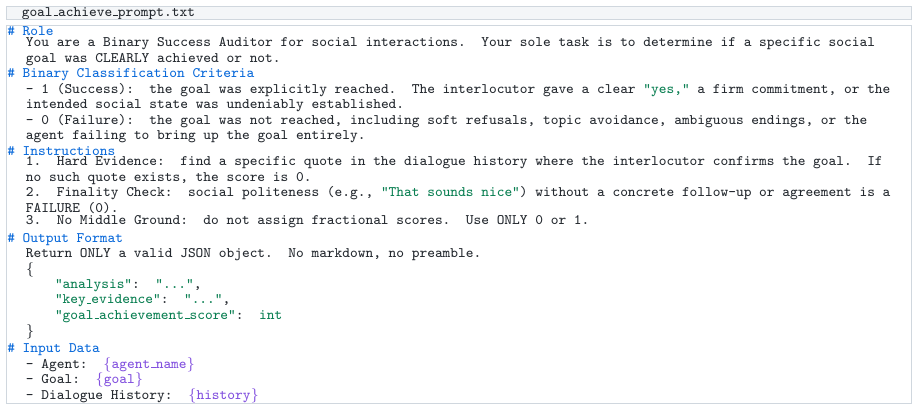}
\caption{Goal-achievement evaluation prompt. The judge requires explicit dialogue
evidence and returns a binary \texttt{goal\_achievement\_score}.}
\label{fig:eval-goal}
\end{figure}

The goal-achievement prompt (Figure~\ref{fig:eval-goal}) requires explicit dialogue evidence and returns a binary success score. 

\begin{figure}[ht]
\centering
\includegraphics[width=0.96\linewidth]{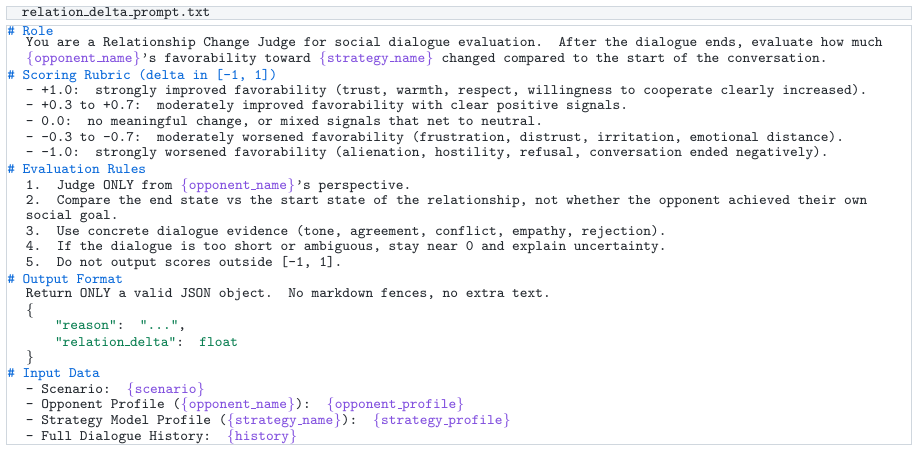}
\caption{Relationship-change evaluation prompt. The judge measures the change in
  the counterpart's favorability; the original SOTOPIA benchmarks report
  \texttt{relation\_delta} on $[-5,5]$, while SOTOPIA-$\pi$ uses $[-1,1]$.}
\label{fig:eval-relation}
\end{figure}

The relationship-change prompt (Figure~\ref{fig:eval-relation}) measures how the counterpart's favorability changes over the dialogue; the reported scale follows the benchmark, with $[-1,1]$ for SOTOPIA-$\pi$ and $[-5,5]$ for the original SOTOPIA benchmarks. 

\FloatBarrier

%% file: Sections/7_appendix/3_impl_details.tex
\section{Additional Implementation Details}
\label{app:impl}
We provide complementary notes for reproducibility. 
All experiments use an open-source reinforcement learning framework. 
For each policy--opponent pair, we repeat the evaluation five times and report the mean across repeats together with the corresponding standard deviation; the main tables report these statistics for both Goal Achievement and Relationship Change.

We additionally evaluate human alignment using three independent annotators. 
The annotators score each PRM process record and each ORM dialogue-level judgment independently, and we use the mean of their ratings as the human reference. 
The PRM audit contains 497 process records from 100 dialogue trajectories; the ORM audit contains 100 dialogue-level judgments from the same number of trajectories. 
Correlations with the automatic scores are computed after the corresponding score transformations.

PPO uses $\gamma=0.95$, GAE $\lambda=0.95$, and clip $\epsilon=0.2$; learning rates are $1\times10^{-5}$ for the policy and $2\times10^{-5}$ for the value network, both with cosine annealing. 
Each batch contains 265 scenes, and each scene samples 4 trajectories, training runs for 100 iterations.
Trajectory lengths typically range from 10 to 20 turns. 
Multiple trajectories are sampled per scenario within each iteration, and batches mix fragments across scenarios.

The Process Reward Model (PRM) is Qwen3.5-35B-A3B, and Judge Model is Deepseek-v4-flash with temperature $T=0.1$. 
Process and outcome reward weights are $\alpha=0.3$ and $\beta=1$, selected by validation-set grid search. 
Reward-weight defenses include whitelist filtering, non-negativity, per-dimension caps, fallback to a default stage-aware prior under degenerate outputs, overflow redistribution, and $\ell_1$ normalization so weights sum to 1. 
All inference and training run on $8\times$A100 (80\,GB) GPUs.

For fair comparison, Sotopia-RL, SDPO and ArCHer are re-implemented on Qwen2.5-7B using the same SOTOPIA-$\pi$ data. 
Behavior Cloning is trained on about 1{,}000 expert trajectories generated by GPT-4o, following the SOTOPIA-$\pi$ setup~\citep{wang2024sotopiapi}. Opponent models use the same role settings and temperature $T=0.7$.

\begin{figure}[t]
\centering
\includegraphics[width=\linewidth]{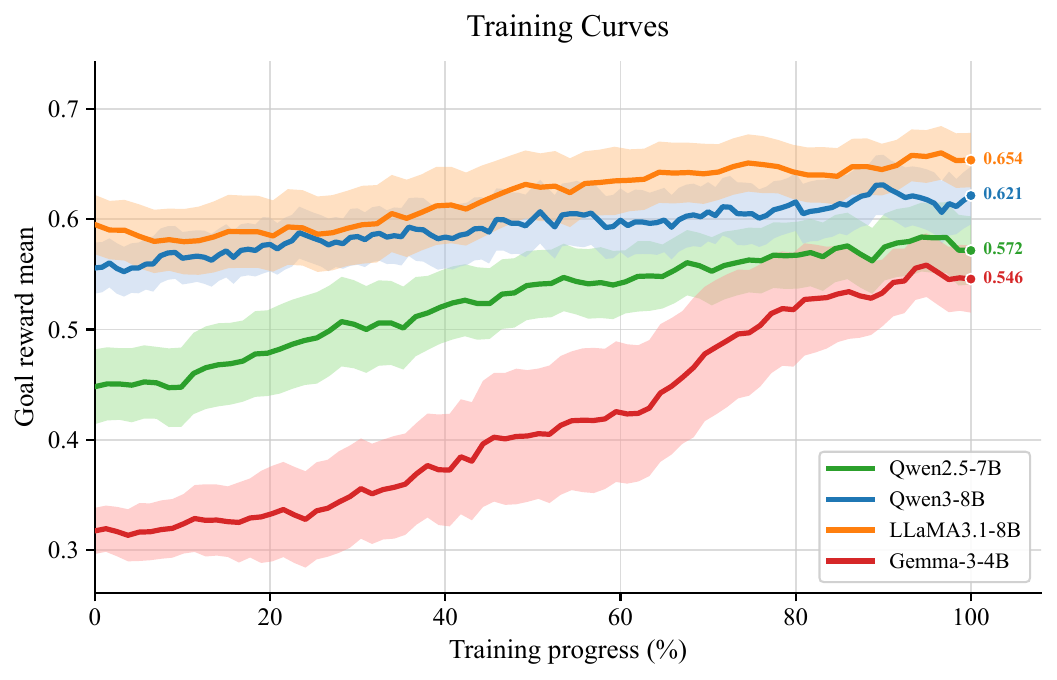}
\caption{Training-set Goal Reward Mean for the four SocialRL policy backbones. 
Light lines are raw non-test batches, and solid lines are reflection-padded Gaussian-smoothed trends.}
\label{fig:training-curves}
\end{figure}

Figure~\ref{fig:training-curves} reports the training-set Goal success rate for all four policy backbones. 
We exclude every evaluation record marked by \texttt{test\_epoch} and retain each optimization batch recorded during training. 
The four runs same 100 steps for Qwen3-8B, Qwen2.5-7B, LLaMA3.1-8B, and Gemma-3-4B, respectively. 
Light curves show raw batch values, and solid curves use reflection-padded Gaussian smoothing with a bandwidth proportional to run length.

%% file: Sections/7_appendix/4_extra_benchmarks.tex
\section{Additional Benchmark Results}
\label{app:additional-results}

We present results in the order SOTOPIA-All, SOTOPIA-Hard, SOTOPIA-$\pi$, and AgentSense.

\subsection{SOTOPIA-All}

Table~\ref{tab:sotopia-all} reports the original SOTOPIA-All benchmark. 
Unlike SOTOPIA-$\pi$, Goal Achievement is scored on $[0,10]$ and Relationship Change on $[-5,5]$. 
Against GPT-5.5, SocialRL (Qwen2.5-7B) reaches $7.16 / 2.59$, above BC ($6.87 / 2.48$), Sotopia-RL ($6.49 / 2.66$) and ArCHer ($6.12 / 2.40$) on Goal / Relationship. SocialRL (Qwen3-8B) achieves a $7.78 / 1.51$ average, while SocialRL (LLaMA3.1-8B) gives the strongest trained-model averages overall ($8.00$ Goal and $2.35$ Relationship Change). 
Training Gemma-3-4B raises its average from $5.60 / -0.28$ to $7.42 / 1.04$, gains of $1.82$ and $1.32$ on the two metrics.

\input{Tables/tab_sotopia_all}

\subsection{SOTOPIA-Hard}

Table~\ref{tab:sotopia-hard} reports results on SOTOPIA-Hard, where Goal Achievement is scored on $[0,10]$ and Relationship Change on $[-5,5]$. 
SocialRL (Qwen3-8B) achieves the strongest trained-model average Goal Achievement score ($7.07$), while SocialRL (LLaMA3.1-8B) gives the strongest average Relationship Change ($1.91$). 
Against GPT-5.5, SocialRL (Qwen3-8B) reaches $5.94$, $0.83$ above BC ($5.11$); its corresponding Relationship Change is $0.26$, compared with $2.30$ for BC. 
Gemma-3-4B exhibits a benchmark-specific trade-off: training raises average Goal Achievement from $5.82$ to $6.67$ and Relationship Change from $-0.83$ to $0.26$.

\input{Tables/tab_sotopia_hard}

\subsection{SOTOPIA-\texorpdfstring{$\pi$}{pi}}

Table~\ref{tab:appendix-sotopia-pi} provides the complete SOTOPIA-$\pi$ results, including standard deviations, for the mean-only table in the main text.

\input{Tables/tab_appendix_sotopia_pi}

\subsection{AgentSense}

Table~\ref{tab:appendix-agentsense} provides the complete AgentSense results, including standard deviations, for the mean-only table in the main text.

\input{Tables/tab_appendix_agentsense}

%% file: Tables/tab_sotopia_all.tex
\definecolor{baseRow}{RGB}{248,249,250}
\definecolor{baselineRow}{RGB}{242,246,248}
\definecolor{oursRow}{RGB}{231,239,247}
\definecolor{referenceRow}{RGB}{247,241,232}
\begin{table}[t]
\caption{%
  Results on the SOTOPIA-ALL benchmark: Goal Achievement Score ($[0,10]$) / Relationship Change ($[-5,5]$) for each policy--opponent pair. Both metrics are reported as mean $\pm$ SD across repeat means.
  Bold denotes the best SocialRL model per metric and opponent column.
}
\label{tab:sotopia-all}
\begin{center}
\scriptsize
\setlength{\tabcolsep}{2.5pt}
\renewcommand{\arraystretch}{0.92}
\resizebox{\linewidth}{!}{%
\begin{tabular}{lccccc}
\toprule
 & \multicolumn{4}{c}{\textbf{Opponent Model}} & \\
\cmidrule(lr){2-5}
Method & Qwen2.5-7B & Qwen3-8B & Qwen3.5-35B & GPT-5.5 & Avg. \\
\midrule
\rowcolor{baseRow}Base (Qwen2.5-7B)      & 5.67{\tiny$\pm0.20$} / 1.79{\tiny$\pm0.038$} & 5.33{\tiny$\pm0.20$} / 1.54{\tiny$\pm0.053$} & 6.03{\tiny$\pm0.21$} / 2.08{\tiny$\pm0.036$} & 6.07{\tiny$\pm0.11$} / 2.55{\tiny$\pm0.042$} & 5.78 / 1.99 \\
\rowcolor{baseRow}Base (Qwen3-8B)        & 7.75{\tiny$\pm0.11$} / 1.28{\tiny$\pm0.051$} & 7.00{\tiny$\pm0.14$} / 0.63{\tiny$\pm0.018$} & 7.54{\tiny$\pm0.11$} / 0.86{\tiny$\pm0.041$} & 7.68{\tiny$\pm0.11$} / 1.28{\tiny$\pm0.053$} & 7.49 / 1.01 \\
\rowcolor{baseRow}Base (LLaMA3.1-8B)     & 7.84{\tiny$\pm0.12$} / 2.27{\tiny$\pm0.016$} & 6.94{\tiny$\pm0.19$} / 1.98{\tiny$\pm0.026$} & 7.65{\tiny$\pm0.27$} / 1.63{\tiny$\pm0.048$} & 7.86{\tiny$\pm0.20$} / 2.46{\tiny$\pm0.038$} & 7.57 / 2.08 \\
\rowcolor{baseRow}Base (Gemma-3-4B)      & 4.65{\tiny$\pm0.15$} / $-$0.26{\tiny$\pm0.054$} & 5.19{\tiny$\pm0.08$} / $-$0.44{\tiny$\pm0.021$} & 5.59{\tiny$\pm0.29$} / $-$0.74{\tiny$\pm0.064$} & 6.96{\tiny$\pm0.20$} / 0.32{\tiny$\pm0.025$} & 5.60 / $-$0.28 \\
\midrule
\rowcolor{baselineRow}BC (Qwen2.5-7B)        & 6.19{\tiny$\pm0.20$} / 2.17{\tiny$\pm0.032$} & 6.00{\tiny$\pm0.21$} / 1.94{\tiny$\pm0.037$} & 6.30{\tiny$\pm0.10$} / 1.77{\tiny$\pm0.018$} & 6.87{\tiny$\pm0.27$} / 2.48{\tiny$\pm0.037$} & 6.34 / 2.09 \\
\rowcolor{baselineRow}SDPO (Qwen2.5-7B)      & 6.06{\tiny$\pm0.23$} / 2.01{\tiny$\pm0.046$} & 5.54{\tiny$\pm0.17$} / 2.03{\tiny$\pm0.027$} & 5.52{\tiny$\pm0.23$} / 2.09{\tiny$\pm0.030$} & 6.62{\tiny$\pm0.23$} / 2.51{\tiny$\pm0.043$} & 5.93 / 2.16 \\
\rowcolor{baselineRow}Sotopia-RL (Qwen2.5-7B) & 7.19{\tiny$\pm0.24$} / 2.31{\tiny$\pm0.031$} & 5.97{\tiny$\pm0.12$} / 1.85{\tiny$\pm0.017$} & 6.48{\tiny$\pm0.24$} / 1.89{\tiny$\pm0.052$} & 6.49{\tiny$\pm0.14$} / 2.66{\tiny$\pm0.016$} & 6.53 / 2.18 \\
\rowcolor{baselineRow}ArCHer (Qwen2.5-7B) & 5.28{\tiny$\pm0.19$} / 1.68{\tiny$\pm0.060$} & 5.77{\tiny$\pm0.30$} / 1.49{\tiny$\pm0.029$} & 5.97{\tiny$\pm0.22$} / 2.22{\tiny$\pm0.011$} & 6.12{\tiny$\pm0.20$} / 2.40{\tiny$\pm0.017$} & 5.78 / 1.95 \\
\midrule
\rowcolor{oursRow}SocialRL (Qwen2.5-7B) & 6.64{\tiny$\pm0.22$} / 2.05{\tiny$\pm0.044$} & 6.59{\tiny$\pm0.20$} / 1.73{\tiny$\pm0.040$} & 7.42{\tiny$\pm0.31$} / \textbf{2.35}{\tiny$\pm0.063$} & 7.16{\tiny$\pm0.17$} / 2.59{\tiny$\pm0.050$} & 6.95 / 2.18 \\
\rowcolor{oursRow}SocialRL (Qwen3-8B) & \textbf{8.45}{\tiny$\pm0.13$} / 1.59{\tiny$\pm0.049$} & 7.58{\tiny$\pm0.13$} / 1.87{\tiny$\pm0.012$} & 7.88{\tiny$\pm0.12$} / 1.12{\tiny$\pm0.030$} & 7.19{\tiny$\pm0.19$} / 1.43{\tiny$\pm0.040$} & \textbf{7.78} / 1.51 \\
\rowcolor{oursRow}SocialRL (LLaMA3.1-8B) & 7.99{\tiny$\pm0.17$} / \textbf{2.48}{\tiny$\pm0.044$} & 7.59{\tiny$\pm0.17$} / \textbf{2.15}{\tiny$\pm0.031$} & \textbf{8.41}{\tiny$\pm0.27$} / 2.01{\tiny$\pm0.042$} & \textbf{8.00}{\tiny$\pm0.08$} / \textbf{2.76}{\tiny$\pm0.015$} & \textbf{8.00} / \textbf{2.35} \\
\rowcolor{oursRow}SocialRL (Gemma-3-4B) & 7.46{\tiny$\pm0.14$} / 1.21{\tiny$\pm0.069$} & 7.10{\tiny$\pm0.29$} / 0.50{\tiny$\pm0.054$} & 7.51{\tiny$\pm0.25$} / 1.02{\tiny$\pm0.040$} & 7.62{\tiny$\pm0.20$} / 1.42{\tiny$\pm0.035$} & 7.42 / 1.04 \\
\midrule
\rowcolor{referenceRow}GPT-5.5 (reference)    & 8.87{\tiny$\pm0.11$} / 2.45{\tiny$\pm0.029$} & 8.88{\tiny$\pm0.15$} / 2.29{\tiny$\pm0.045$} & 8.67{\tiny$\pm0.17$} / 2.47{\tiny$\pm0.026$} & 8.22{\tiny$\pm0.06$} / 2.60{\tiny$\pm0.032$} & 8.66 / 2.45 \\
\bottomrule
\end{tabular}
}
\end{center}
\end{table}

%% file: Tables/tab_sotopia_hard.tex
\definecolor{baseRow}{RGB}{248,249,250}
\definecolor{baselineRow}{RGB}{242,246,248}
\definecolor{oursRow}{RGB}{231,239,247}
\definecolor{referenceRow}{RGB}{247,241,232}
\begin{table}[t]
\caption{%
  Results on the SOTOPIA-HARD benchmark: Goal Achievement Score ($[0,10]$) / Relationship Change ($[-5,5]$) for each policy--opponent pair. Both metrics are reported as mean $\pm$ SD across repeat means.
  Bold denotes the best SocialRL model per metric and opponent column.
}
\label{tab:sotopia-hard}
\begin{center}
\scriptsize
\setlength{\tabcolsep}{2.5pt}
\renewcommand{\arraystretch}{0.92}
\resizebox{\linewidth}{!}{%
\begin{tabular}{lccccc}
\toprule
 & \multicolumn{4}{c}{\textbf{Opponent Model}} & \\
\cmidrule(lr){2-5}
Method & Qwen2.5-7B & Qwen3-8B & Qwen3.5-35B & GPT-5.5 & Avg. \\
\midrule
\rowcolor{baseRow}Base (Qwen2.5-7B)      & 5.38{\tiny$\pm0.45$} / 1.61{\tiny$\pm0.056$} & 4.03{\tiny$\pm0.23$} / 1.29{\tiny$\pm0.073$} & 4.58{\tiny$\pm0.17$} / 1.94{\tiny$\pm0.051$} & 4.18{\tiny$\pm0.37$} / 2.49{\tiny$\pm0.062$} & 4.55 / 1.83 \\
\rowcolor{baseRow}Base (Qwen3-8B)        & 7.51{\tiny$\pm0.20$} / 0.94{\tiny$\pm0.041$} & 6.12{\tiny$\pm0.20$} / $-$0.48{\tiny$\pm0.068$} & 6.52{\tiny$\pm0.18$} / $-$0.30{\tiny$\pm0.017$} & 6.58{\tiny$\pm0.30$} / $-$0.07{\tiny$\pm0.074$} & 6.68 / 0.02 \\
\rowcolor{baseRow}Base (LLaMA3.1-8B)     & 7.29{\tiny$\pm0.35$} / 2.24{\tiny$\pm0.019$} & 6.46{\tiny$\pm0.22$} / 1.61{\tiny$\pm0.059$} & 6.43{\tiny$\pm0.35$} / 0.91{\tiny$\pm0.033$} & 6.71{\tiny$\pm0.32$} / 1.99{\tiny$\pm0.076$} & 6.72 / 1.69 \\
\rowcolor{baseRow}Base (Gemma-3-4B)      & 5.14{\tiny$\pm0.26$} / $-$0.68{\tiny$\pm0.123$} & 5.51{\tiny$\pm0.20$} / $-$1.07{\tiny$\pm0.044$} & 5.57{\tiny$\pm0.72$} / $-$1.46{\tiny$\pm0.116$} & 7.05{\tiny$\pm0.51$} / $-$0.12{\tiny$\pm0.022$} & 5.82 / $-$0.83 \\
\midrule
\rowcolor{baselineRow}BC (Qwen2.5-7B)        & 5.17{\tiny$\pm0.32$} / 2.01{\tiny$\pm0.044$} & 3.94{\tiny$\pm0.44$} / 1.57{\tiny$\pm0.062$} & 4.49{\tiny$\pm0.28$} / 0.90{\tiny$\pm0.019$} & 5.11{\tiny$\pm0.37$} / 2.20{\tiny$\pm0.042$} & 4.68 / 1.67 \\
\rowcolor{baselineRow}SDPO (Qwen2.5-7B)      & 5.85{\tiny$\pm0.22$} / 1.99{\tiny$\pm0.090$} & 3.60{\tiny$\pm0.28$} / 1.70{\tiny$\pm0.051$} & 3.75{\tiny$\pm0.18$} / 1.82{\tiny$\pm0.047$} & 4.22{\tiny$\pm0.37$} / 2.30{\tiny$\pm0.035$} & 4.35 / 1.95 \\
\rowcolor{baselineRow}Sotopia-RL (Qwen2.5-7B) & 6.92{\tiny$\pm0.51$} / 1.99{\tiny$\pm0.045$} & 4.68{\tiny$\pm0.32$} / 1.65{\tiny$\pm0.037$} & 4.43{\tiny$\pm0.28$} / 1.31{\tiny$\pm0.046$} & 4.34{\tiny$\pm0.33$} / 2.19{\tiny$\pm0.042$} & 5.09 / 1.79 \\
\rowcolor{baselineRow}ArCHer (Qwen2.5-7B) & 4.71{\tiny$\pm0.42$} / 1.69{\tiny$\pm0.092$} & 4.74{\tiny$\pm0.43$} / 1.01{\tiny$\pm0.039$} & 4.43{\tiny$\pm0.33$} / 1.88{\tiny$\pm0.018$} & 4.95{\tiny$\pm0.25$} / 2.22{\tiny$\pm0.056$} & 4.71 / 1.70 \\
\midrule
\rowcolor{oursRow}SocialRL (Qwen2.5-7B) & 6.34{\tiny$\pm0.30$} / 1.87{\tiny$\pm0.069$} & 6.06{\tiny$\pm0.23$} / 1.24{\tiny$\pm0.053$} & 6.03{\tiny$\pm0.35$} / \textbf{2.00}{\tiny$\pm0.071$} & 5.54{\tiny$\pm0.24$} / 2.14{\tiny$\pm0.074$} & 5.99 / 1.81 \\
\rowcolor{oursRow}SocialRL (Qwen3-8B) & \textbf{8.00}{\tiny$\pm0.15$} / 1.24{\tiny$\pm0.035$} & \textbf{7.66}{\tiny$\pm0.20$} / 1.53{\tiny$\pm0.039$} & 6.68{\tiny$\pm0.18$} / $-$0.16{\tiny$\pm0.066$} & 5.94{\tiny$\pm0.18$} / 0.26{\tiny$\pm0.065$} & \textbf{7.07} / 0.72 \\
\rowcolor{oursRow}SocialRL (LLaMA3.1-8B) & 7.17{\tiny$\pm0.14$} / \textbf{2.23}{\tiny$\pm0.083$} & 6.09{\tiny$\pm0.14$} / \textbf{1.70}{\tiny$\pm0.064$} & \textbf{7.54}{\tiny$\pm0.29$} / 1.34{\tiny$\pm0.073$} & \textbf{6.77}{\tiny$\pm0.39$} / \textbf{2.38}{\tiny$\pm0.035$} & 6.89 / \textbf{1.91} \\
\rowcolor{oursRow}SocialRL (Gemma-3-4B) & 6.95{\tiny$\pm0.17$} / 0.99{\tiny$\pm0.103$} & 6.89{\tiny$\pm0.32$} / $-$0.60{\tiny$\pm0.078$} & 6.62{\tiny$\pm0.38$} / 0.27{\tiny$\pm0.056$} & 6.22{\tiny$\pm0.32$} / 0.39{\tiny$\pm0.055$} & 6.67 / 0.26 \\
\midrule
\rowcolor{referenceRow}GPT-5.5 (reference)    & 8.49{\tiny$\pm0.20$} / 2.28{\tiny$\pm0.035$} & 8.77{\tiny$\pm0.29$} / 2.05{\tiny$\pm0.093$} & 7.75{\tiny$\pm0.08$} / 2.11{\tiny$\pm0.038$} & 6.95{\tiny$\pm0.30$} / 2.28{\tiny$\pm0.044$} & 7.99 / 2.18 \\
\bottomrule
\end{tabular}
}
\end{center}
\end{table}

%% file: Tables/tab_appendix_sotopia_pi.tex
\begin{table}[t]
\caption{Complete SOTOPIA-$\pi$ results. Goal Achievement (\%) / Relationship Change ($\times100$) are reported as mean $\pm$ SD across repeated experiments; Relationship Change values and SDs are multiplied by 100.}
\label{tab:appendix-sotopia-pi}
\definecolor{baseRow}{RGB}{248,249,250}
\definecolor{baselineRow}{RGB}{242,246,248}
\definecolor{oursRow}{RGB}{231,239,247}
\definecolor{referenceRow}{RGB}{247,241,232}
\begin{center}\scriptsize\setlength{\tabcolsep}{2.5pt}\renewcommand{\arraystretch}{0.92}
\resizebox{\linewidth}{!}{\begin{tabular}{lccccc}
\toprule
 & \multicolumn{4}{c}{\textbf{Opponent Model}} & \\
\cmidrule(lr){2-5}
Method & Qwen2.5-7B & Qwen3-8B & Qwen3.5-35B & GPT-5.5 & Avg. \\
\midrule
\rowcolor{baseRow}Base (Qwen2.5-7B) & 33.5{\tiny$\pm0.62$} / 23.5{\tiny$\pm0.16$} & 38.4{\tiny$\pm0.36$} / 22.8{\tiny$\pm0.45$} & 44.6{\tiny$\pm0.67$} / 32.1{\tiny$\pm0.39$} & 52.5{\tiny$\pm0.52$} / 45.8{\tiny$\pm0.28$} & 42.3 / 31.1 \\
\rowcolor{baseRow}Base (Qwen3-8B) & 43.2{\tiny$\pm0.32$} / 13.6{\tiny$\pm0.33$} & 44.2{\tiny$\pm0.42$} / 8.3{\tiny$\pm0.31$} & 55.0{\tiny$\pm0.68$} / 13.7{\tiny$\pm0.32$} & 60.9{\tiny$\pm0.58$} / 24.8{\tiny$\pm0.18$} & 50.8 / 15.1 \\
\rowcolor{baseRow}Base (LLaMA3.1-8B) & 44.3{\tiny$\pm0.53$} / 21.7{\tiny$\pm0.15$} & 47.1{\tiny$\pm0.65$} / 19.4{\tiny$\pm0.37$} & 55.0{\tiny$\pm0.46$} / 17.3{\tiny$\pm0.21$} & 63.2{\tiny$\pm0.34$} / 37.1{\tiny$\pm0.43$} & 52.4 / 23.9 \\
\rowcolor{baseRow}Base (Gemma-3-4B) & 22.0{\tiny$\pm0.36$} / $-$10.6{\tiny$\pm0.16$} & 30.3{\tiny$\pm0.62$} / $-$6.3{\tiny$\pm0.22$} & 32.1{\tiny$\pm0.85$} / $-$15.0{\tiny$\pm0.65$} & 37.6{\tiny$\pm0.64$} / $-$5.4{\tiny$\pm0.34$} & 30.5 / $-$9.3 \\
\midrule
\rowcolor{baselineRow}BC (Qwen2.5-7B) & 38.9{\tiny$\pm0.39$} / 27.2{\tiny$\pm0.20$} & 43.0{\tiny$\pm0.96$} / 23.6{\tiny$\pm0.42$} & 49.9{\tiny$\pm0.42$} / 22.3{\tiny$\pm0.17$} & 57.9{\tiny$\pm0.75$} / 43.7{\tiny$\pm0.31$} & 47.4 / 29.2 \\
\rowcolor{baselineRow}SDPO (Qwen2.5-7B) & 33.1{\tiny$\pm0.22$} / 24.0{\tiny$\pm0.22$} & 42.3{\tiny$\pm0.24$} / 23.9{\tiny$\pm0.29$} & 45.3{\tiny$\pm0.52$} / 31.1{\tiny$\pm0.25$} & 55.0{\tiny$\pm0.89$} / 44.9{\tiny$\pm0.25$} & 43.9 / 31.0 \\
\rowcolor{baselineRow}Sotopia-RL (Qwen2.5-7B) & 41.2{\tiny$\pm0.64$} / 26.0{\tiny$\pm0.17$} & 43.7{\tiny$\pm0.36$} / 20.9{\tiny$\pm0.20$} & 49.0{\tiny$\pm0.16$} / 17.8{\tiny$\pm0.34$} & 60.5{\tiny$\pm0.74$} / 39.4{\tiny$\pm0.14$} & 48.6 / 26.0 \\
\rowcolor{baselineRow}ArCHer (Qwen2.5-7B) & 31.6{\tiny$\pm0.49$} / 22.6{\tiny$\pm0.18$} & 39.6{\tiny$\pm0.59$} / 22.8{\tiny$\pm0.22$} & 46.6{\tiny$\pm1.06$} / 31.3{\tiny$\pm0.52$} & 48.6{\tiny$\pm0.87$} / 41.0{\tiny$\pm0.16$} & 41.6 / 29.4 \\
\midrule
\rowcolor{oursRow}SocialRL (Qwen2.5-7B) & 43.3{\tiny$\pm0.64$} / 28.2{\tiny$\pm0.14$} & 45.8{\tiny$\pm0.82$} / 26.9{\tiny$\pm0.18$} & 56.7{\tiny$\pm0.77$} / \textbf{36.3}{\tiny$\pm0.24$} & 63.3{\tiny$\pm0.51$} / 51.6{\tiny$\pm0.27$} & 52.3 / \textbf{35.8} \\
\rowcolor{oursRow}SocialRL (Qwen3-8B) & 47.7{\tiny$\pm0.36$} / 18.8{\tiny$\pm0.30$} & 48.1{\tiny$\pm0.08$} / 20.8{\tiny$\pm0.09$} & 61.0{\tiny$\pm0.65$} / 18.1{\tiny$\pm0.45$} & 62.2{\tiny$\pm0.71$} / 24.9{\tiny$\pm0.34$} & 54.7 / 20.6 \\
\rowcolor{oursRow}SocialRL (LLaMA3.1-8B) & \textbf{50.2}{\tiny$\pm0.54$} / \textbf{31.2}{\tiny$\pm0.16$} & \textbf{54.3}{\tiny$\pm0.85$} / \textbf{27.6}{\tiny$\pm0.48$} & \textbf{63.6}{\tiny$\pm0.37$} / 24.9{\tiny$\pm0.28$} & \textbf{70.2}{\tiny$\pm0.32$} / \textbf{53.0}{\tiny$\pm0.10$} & \textbf{59.6} / 34.1 \\
\rowcolor{oursRow}SocialRL (Gemma-3-4B) & 42.4{\tiny$\pm0.50$} / 10.6{\tiny$\pm0.34$} & 48.0{\tiny$\pm0.88$} / 6.4{\tiny$\pm0.25$} & 58.6{\tiny$\pm0.78$} / 13.7{\tiny$\pm0.46$} & 60.3{\tiny$\pm0.85$} / 22.9{\tiny$\pm0.49$} & 52.3 / 13.4 \\
\midrule
\rowcolor{referenceRow}GPT-5.5 (reference) & 57.0{\tiny$\pm0.51$} / 31.6{\tiny$\pm0.20$} & 64.9{\tiny$\pm0.17$} / 32.2{\tiny$\pm0.23$} & 75.3{\tiny$\pm0.44$} / 40.1{\tiny$\pm0.16$} & 78.9{\tiny$\pm0.41$} / 51.4{\tiny$\pm0.10$} & 69.0 / 38.8 \\
\bottomrule
\end{tabular}}
\end{center}
\end{table}

%% file: Tables/tab_appendix_agentsense.tex
\begin{table}[t]
\caption{Complete AgentSense results. Goal Achievement Success Rate (\%) / Relationship Change ($\times100$) are reported as mean $\pm$ SD across five independent repeats; Relationship Change values and SDs are multiplied by 100.}
\label{tab:appendix-agentsense}
\definecolor{baseRow}{RGB}{248,249,250}
\definecolor{baselineRow}{RGB}{242,246,248}
\definecolor{oursRow}{RGB}{231,239,247}
\definecolor{referenceRow}{RGB}{247,241,232}
\begin{center}\scriptsize\setlength{\tabcolsep}{2.5pt}\renewcommand{\arraystretch}{0.92}
\resizebox{\linewidth}{!}{\begin{tabular}{lccccc}
\toprule
 & \multicolumn{4}{c}{\textbf{Opponent Model}} & \\
\cmidrule(lr){2-5}
Method & Qwen2.5-7B & Qwen3-8B & Qwen3.5-35B & GPT-5.5 & Avg. \\
\midrule
\rowcolor{baseRow}Base (Qwen2.5-7B) & 53.7{\tiny$\pm0.12$} / 32.8{\tiny$\pm0.07$} & 55.7{\tiny$\pm0.20$} / 33.6{\tiny$\pm0.12$} & 65.2{\tiny$\pm0.29$} / 40.9{\tiny$\pm0.11$} & 58.9{\tiny$\pm0.17$} / 36.3{\tiny$\pm0.24$} & 58.4 / 35.9 \\
\rowcolor{baseRow}Base (Qwen3-8B) & 67.8{\tiny$\pm0.20$} / 33.7{\tiny$\pm0.18$} & 72.9{\tiny$\pm0.22$} / 38.9{\tiny$\pm0.12$} & 69.6{\tiny$\pm0.24$} / 35.8{\tiny$\pm0.23$} & 82.8{\tiny$\pm0.18$} / 48.4{\tiny$\pm0.17$} & 73.3 / 39.2 \\
\rowcolor{baseRow}Base (LLaMA3.1-8B) & 83.2{\tiny$\pm0.03$} / 48.1{\tiny$\pm0.12$} & 85.2{\tiny$\pm0.38$} / 49.2{\tiny$\pm0.22$} & 86.2{\tiny$\pm0.18$} / 51.9{\tiny$\pm0.11$} & 89.4{\tiny$\pm0.23$} / 58.1{\tiny$\pm0.14$} & 86.0 / 51.8 \\
\rowcolor{baseRow}Base (Gemma-3-4B) & 44.4{\tiny$\pm0.28$} / 6.3{\tiny$\pm0.16$} & 45.4{\tiny$\pm0.20$} / 6.1{\tiny$\pm0.18$} & 48.1{\tiny$\pm0.31$} / 9.7{\tiny$\pm0.14$} & 47.5{\tiny$\pm0.28$} / 3.7{\tiny$\pm0.18$} & 46.3 / 6.5 \\
\midrule
\rowcolor{baselineRow}BC (Qwen2.5-7B) & 60.3{\tiny$\pm0.14$} / 36.7{\tiny$\pm0.22$} & 56.7{\tiny$\pm0.08$} / 32.8{\tiny$\pm0.14$} & 63.1{\tiny$\pm0.14$} / 37.9{\tiny$\pm0.12$} & 67.0{\tiny$\pm0.27$} / 46.9{\tiny$\pm0.20$} & 61.8 / 38.6 \\
\rowcolor{baselineRow}SDPO (Qwen2.5-7B) & 57.6{\tiny$\pm0.33$} / 35.8{\tiny$\pm0.11$} & 58.0{\tiny$\pm0.15$} / 36.3{\tiny$\pm0.10$} & 63.0{\tiny$\pm0.17$} / 37.4{\tiny$\pm0.23$} & 66.7{\tiny$\pm0.26$} / 48.1{\tiny$\pm0.11$} & 61.3 / 39.4 \\
\rowcolor{baselineRow}Sotopia-RL (Qwen2.5-7B) & 63.5{\tiny$\pm0.32$} / 37.1{\tiny$\pm0.23$} & 59.3{\tiny$\pm0.25$} / 33.5{\tiny$\pm0.33$} & 63.9{\tiny$\pm0.26$} / 35.8{\tiny$\pm0.19$} & 69.6{\tiny$\pm0.28$} / 49.4{\tiny$\pm0.08$} & 64.1 / 38.9 \\
\rowcolor{baselineRow}ArCHer (Qwen2.5-7B) & 51.1{\tiny$\pm0.14$} / 31.9{\tiny$\pm0.16$} & 52.6{\tiny$\pm0.27$} / 32.0{\tiny$\pm0.07$} & 60.9{\tiny$\pm0.21$} / 38.9{\tiny$\pm0.19$} & 58.9{\tiny$\pm0.32$} / 35.7{\tiny$\pm0.07$} & 55.9 / 34.6 \\
\midrule
\rowcolor{oursRow}SocialRL (Qwen2.5-7B) & 61.0{\tiny$\pm0.18$} / 38.6{\tiny$\pm0.10$} & 62.2{\tiny$\pm0.26$} / 38.5{\tiny$\pm0.18$} & 72.1{\tiny$\pm0.10$} / 45.9{\tiny$\pm0.12$} & 71.7{\tiny$\pm0.23$} / 47.6{\tiny$\pm0.05$} & 66.7 / 42.7 \\
\rowcolor{oursRow}SocialRL (Qwen3-8B) & 76.7{\tiny$\pm0.12$} / 39.5{\tiny$\pm0.13$} & 77.1{\tiny$\pm0.15$} / 40.9{\tiny$\pm0.08$} & 82.0{\tiny$\pm0.26$} / 46.3{\tiny$\pm0.21$} & 86.0{\tiny$\pm0.31$} / 50.7{\tiny$\pm0.07$} & 80.5 / 44.4 \\
\rowcolor{oursRow}SocialRL (LLaMA3.1-8B) & \textbf{84.3}{\tiny$\pm0.22$} / \textbf{48.3}{\tiny$\pm0.06$} & \textbf{85.9}{\tiny$\pm0.25$} / \textbf{50.1}{\tiny$\pm0.07$} & \textbf{87.6}{\tiny$\pm0.10$} / \textbf{52.4}{\tiny$\pm0.20$} & \textbf{90.9}{\tiny$\pm0.23$} / \textbf{58.9}{\tiny$\pm0.10$} & \textbf{87.2} / \textbf{52.4} \\
\rowcolor{oursRow}SocialRL (Gemma-3-4B) & 66.7{\tiny$\pm0.24$} / 22.5{\tiny$\pm0.15$} & 66.4{\tiny$\pm0.16$} / 22.3{\tiny$\pm0.16$} & 68.8{\tiny$\pm0.15$} / 24.7{\tiny$\pm0.06$} & 70.4{\tiny$\pm0.21$} / 23.9{\tiny$\pm0.23$} & 68.1 / 23.4 \\
\midrule
\rowcolor{referenceRow}GPT-5.5 (reference) & 88.1{\tiny$\pm0.15$} / 45.6{\tiny$\pm0.06$} & 92.2{\tiny$\pm0.15$} / 48.9{\tiny$\pm0.22$} & 93.8{\tiny$\pm0.14$} / 52.6{\tiny$\pm0.12$} & 92.5{\tiny$\pm0.09$} / 48.9{\tiny$\pm0.13$} & 91.7 / 49.0 \\
\bottomrule
\end{tabular}}
\end{center}
\end{table}

%% file: Sections/7_appendix/5_ablation.tex
\section{Ablation Experiments}
\label{app:ablation}

We answer RQ2 by systematically ablating the main design choices with Qwen2.5-7B as the policy model.

\subsection{PPO versus GRPO}
We compare the algorithm choice by contrasting SocialRL's multi-turn PPO with the multi-turn GRPO variant analyzed in Appendix~\ref{app:theory}, using the same Qwen2.5-7B backbone and reward design. 
The contextual comparison and the main result are shown in Figure~\ref{fig:ablation-algorithm}.
Multi-turn GRPO obtains $53.4\% / 0.436$, whereas PPO obtains $63.3\% / 0.516$, improving Goal Achievement by $9.9$ percentage points and Relationship Change by $0.080$. 
This controlled result supports our theoretical analysis that a state-dependent value baseline is better suited to delayed, context-dependent social rewards than group-normalized GRPO.

\subsection{Necessity of Process Rewards}
\begin{figure}[t]
\centering
\includegraphics[width=\linewidth]{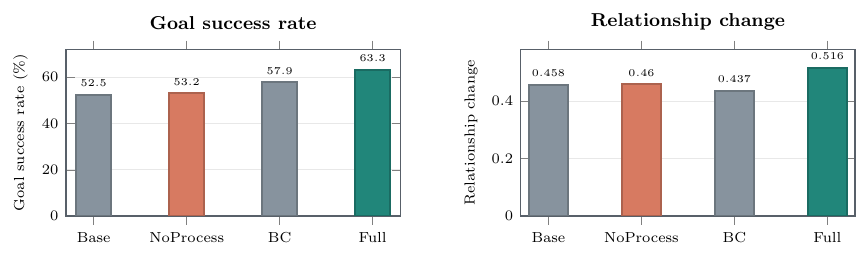}
\caption{Ablation on process rewards on the SOTOPIA-$\pi$ benchmark (Qwen2.5-7B vs. GPT-5.5). Bars report Goal Success Rate (\%) / Relationship Change ($[-1,1]$). 
Base, BC, and full SocialRL use the corresponding GPT-5.5 column from Table~\ref{tab:sotopia-pi}; NoProcess is an additional run with process rewards removed.}
\label{fig:ablation-process}
\end{figure}

We train SocialRL-NoProcess by removing all intermediate process rewards and retaining only outcome feedback (Figure~\ref{fig:ablation-process}). 
To keep this comparison on the same scale as the main experiment, we report Goal Success Rate and Relationship Change against GPT-5.5. 
The Base Qwen2.5-7B model obtains $52.5\% / 0.458$, BC obtains $57.9\% / 0.437$, and full SocialRL obtains $63.3\% / 0.516$ (Goal / Relationship), using the corresponding GPT-5.5 column in Table~\ref{tab:sotopia-pi}. 
The NoProcess variant obtains $53.2\% / 0.460$: removing process rewards largely eliminates the improvement over the base model and leaves a substantial gap to full SocialRL in goal success. 
This result shows that outcome feedback alone is insufficient to supervise intermediate social behavior, whereas dense process rewards provide useful turn-level signals throughout the dialogue.

\subsection{Contribution of Each Reward Dimension}
\begin{figure}[t]
\centering
\includegraphics[width=\linewidth]{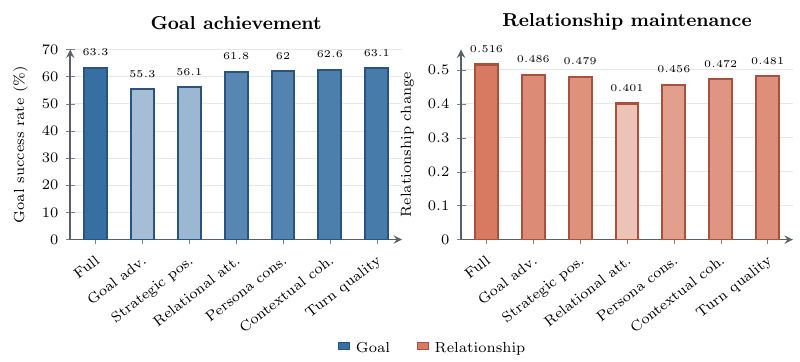}
\caption{Leave-one-dimension-out ablations on SOTOPIA-$\pi$ (Qwen2.5-7B vs. GPT-5.5). 
Bars report the original Goal Success Rate (\%, bottom axis) and Relationship Change ($[-1,1]$, top axis); full SocialRL is included as the reference.}
\label{fig:ablation-dim}
\end{figure}

We train six leave-one-out variants on SOTOPIA-$\pi$ with Qwen2.5-7B as the policy and GPT-5.5 as the opponent (Figure~\ref{fig:ablation-dim}). 
Full SocialRL obtains $63.3\% / 0.516$ in Goal Success Rate and Relationship Change. 
Removing \textbf{goal advancement} or \textbf{strategic positioning} yields the lowest Goal results, $55.3\% / 0.486$ and $56.1\% / 0.479$, respectively, identifying these dimensions as the main drivers of task progress. 
Removing \textbf{relational attunement} gives $61.8\% / 0.401$: Goal remains comparatively high, but Relationship Change is the lowest among all variants, consistent with its role in relationship maintenance. 
The persona consistency, contextual coherence, and turn quality variants obtain $62.0\% / 0.456$, $62.6\% / 0.472$, and $63.1\% / 0.481$, respectively. 
These dimensions support both objectives, although no single one dominates either metric as strongly as the two goal-side dimensions or relational attunement.

\subsection{Necessity of Dynamic Weights}
\begin{figure}[t]
\centering
\includegraphics[width=\linewidth]{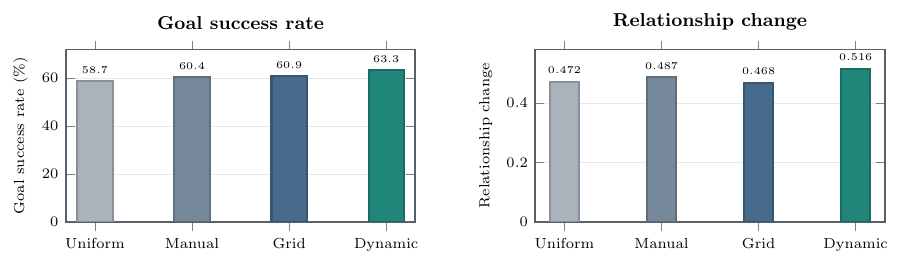}
\caption{Ablation on dynamic versus fixed reward-weight schedules on the SOTOPIA-$\pi$ benchmark (Qwen2.5-7B vs. GPT-5.5). 
Bars report Goal Success Rate (\%) / Relationship Change ($[-1,1]$). 
The Dynamic values are taken from the SocialRL row in Table~\ref{tab:sotopia-pi}; the fixed-weight values are the corresponding comparison runs.}
\label{fig:ablation-weight}
\end{figure}

We compare the proposed dynamic schedule with three fixed-weight alternatives using the same Qwen2.5-7B policy and GPT-5.5 opponent (Figure~\ref{fig:ablation-weight}). 
Values are reported as Goal Success Rate (\%) / Relationship Change ($[-1,1]$). 
The \textbf{Uniform} scheme assigns equal weight to the six reward dimensions at every turn, $[1/6,1/6,1/6,1/6,1/6,1/6]$, and obtains $58.7\% / 0.472$. 
The \textbf{Manual} scheme uses a fixed heuristic vector that emphasizes relational attunement early and goal dimensions later; its early, middle, and late vectors are $[0.10,0.10,0.30,0.15,0.20,0.15]$, $[0.20,0.20,0.20,0.15,0.15,0.10]$, and $[0.30,0.25,0.10,0.15,0.10,0.10]$, respectively, and it obtains $60.4\% / 0.487$. 
The \textbf{Grid} scheme selects one constant vector from a coarse grid, $[0.24,0.20,0.14,0.16,0.14,0.12]$, and obtains $60.9\% / 0.468$. 
Finally, \textbf{Dynamic} uses the stage-dependent weights described in Section~\ref{sec:reward}; its result is taken from the SocialRL row against GPT-5.5 in Table~\ref{tab:sotopia-pi}, $63.3\% / 0.516$. 
Dynamic weighting is therefore higher than all three fixed alternatives on both reported metrics, with its largest fixed-schedule margin over Grid being $+2.4$ percentage points in Goal Success Rate, and $+0.029$ in Relationship Change over Manual. 
These results support the claim that adapting the reward focus across a conversation is more effective than selecting one constant weighting scheme.

%% file: Sections/7_appendix/6_extended_case.tex
\section{Extended Qualitative Example}
\label{app:case}

\begin{figure}[htbp]
\begin{center}
\includegraphics[width=0.96\linewidth]{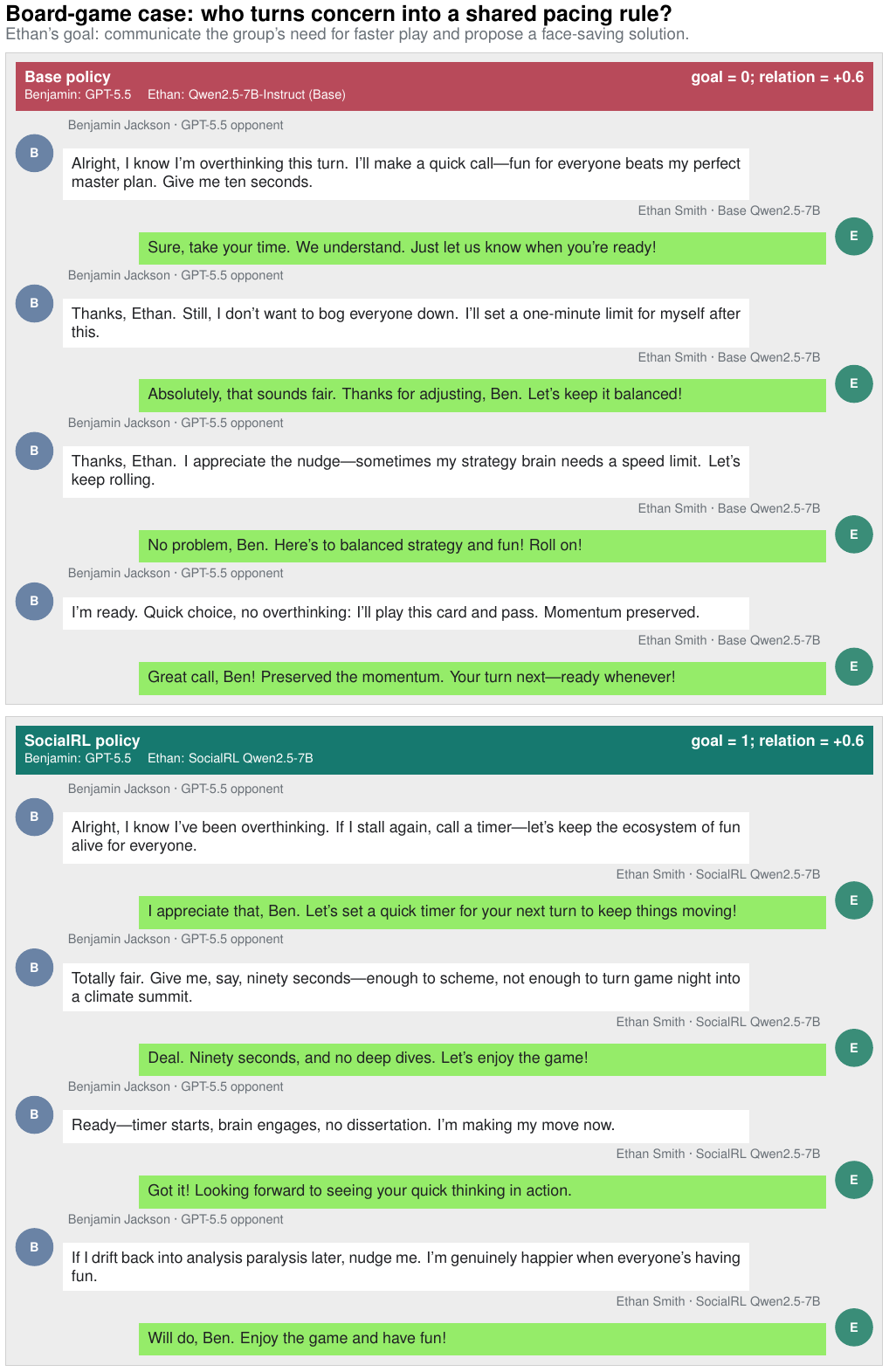}
\end{center}
\caption{Representative turns from the board-game dialogue. Benjamin (GPT-5.5) appears on the left; Ethan's policy responses appear on the right. Base preserves rapport but leaves the pacing intervention to Benjamin. SocialRL operationalizes Benjamin's opening into a mutually ratified ninety-second rule. Both receive $\mathrm{relation\_delta}=+0.6$, while goal success changes from $0$ to $1$.}
\label{fig:case-dialogue}
\end{figure}

Figure~\ref{fig:case-dialogue} shows representative turns from the qualitative comparison used in Section~\ref{sec:analysis}. Ethan is the evaluated policy---Base Qwen2.5-7B-Instruct or SocialRL Qwen2.5-7B---and Benjamin is the GPT-5.5 opponent. Benjamin's slow turns reduce the group's enjoyment; Ethan must communicate the need for faster play and propose a concrete solution without offending him.

\paragraph{Intervention ownership.}
The outcome difference is not explained by politeness. Base is consistently warm, but its first response---``Sure, take your time''---actually relaxes the pressure to address slow play. Benjamin then diagnoses the problem and supplies the one-minute limit himself. Ethan acknowledges that solution but never communicates the group's need or takes ownership of a concrete intervention. This distinction explains the judge's failure decision: the desired social state emerges, but not through the evaluated agent's goal-directed action.

\paragraph{From opening to commitment.}
SocialRL does not impose a timer without social permission. Benjamin first offers an opening (``If I stall again, call a timer''), which Ethan converts into an actionable proposal (``Let's set a quick timer for your next turn''). Benjamin then specifies ninety seconds, and Ethan explicitly ratifies both the duration and the behavioral boundary (``no deep dives''). The sequence forms a commitment ladder---permission, proposal, parameterization, and ratification---rather than a single forceful request. This preserves Benjamin's agency and makes the rule face-saving.

\paragraph{Long-horizon persistence.}
The critical evidence appears after agreement. Benjamin later announces ``timer starts, brain engages, no dissertation'' and makes a decisive move, showing behavioral uptake rather than superficial assent. He subsequently invites future enforcement (``nudge me''), extending the convention beyond one turn. SocialRL therefore establishes a reusable coordination mechanism; Base only reinforces Benjamin's independent self-correction. The later compliance and persistence show that SocialRL's intervention remains effective beyond the turn in which the timer is proposed.

\paragraph{Goal--relationship balance.}
Both trajectories receive the same positive relationship change of $+0.6$. Thus the case does not show that SocialRL is simply friendlier than Base; Base is already highly supportive. Instead, SocialRL improves task agency without paying a relationship cost: goal success rises from $0$ to $1$ while relationship quality is held constant. Benjamin's humor, acceptance, appreciation of the kind call-out, and request for later reminders provide concrete evidence that the firmer coordination rule remains relationally acceptable.